\documentclass{article} 
\usepackage{iclr2022_conference,times}

\usepackage{amsmath,amsfonts,bm}

\def\eqref#1{equation~\ref{#1}}

\def\1{\bm{1}}

\def\vt{{\bm{t}}}

\def\mA{{\bm{A}}}
\def\mB{{\bm{B}}}

\def\mF{{\bm{F}}}

\def\mH{{\bm{H}}}

\def\mK{{\bm{K}}}

\def\mS{{\bm{S}}}
\def\mT{{\bm{T}}}

\def\mX{{\bm{X}}}

\def\mZ{{\bm{Z}}}

\DeclareMathAlphabet{\mathsfit}{\encodingdefault}{\sfdefault}{m}{sl}
\SetMathAlphabet{\mathsfit}{bold}{\encodingdefault}{\sfdefault}{bx}{n}

\newcommand{\KL}{D_{\mathrm{KL}}}
\newcommand{\Var}{\mathrm{Var}}

\newcommand{\Cov}{\mathrm{Cov}}

\DeclareMathOperator*{\argmax}{arg\,max}
\DeclareMathOperator*{\argmin}{arg\,min}

\usepackage[T1]{fontenc}
\usepackage[utf8]{inputenc}
\usepackage{algorithm,algpseudocode}
\usepackage{amsmath}
\usepackage{amsthm}
\usepackage{graphicx}
\usepackage{xargs}[2008/03/08]
\usepackage{microtype}

\usepackage{authblk}
\makeatletter
\renewcommand\maketitle{\AB@maketitle} 
\renewcommand\AB@affilsepx{\quad\protect\Affilfont} 
\makeatother
\renewcommand\Affilfont{\small} 
\usepackage{wrapfig}

\usepackage[unicode=true,
 bookmarks=false,
 breaklinks=false,pdfborder={0 0 1},backref=section,colorlinks=false]
 {hyperref}

\makeatletter

\providecommand{\tabularnewline}{\\}
\newcommand{\lyxdot}{.}

\theoremstyle{plain}
\newtheorem{thm}{\protect\theoremname}
\ifx\proof\undefined
\newenvironment{proof}[1][\protect\proofname]{\par
	\normalfont\topsep6\p@\@plus6\p@\relax
	\trivlist
	\itemindent\parindent
	\item[\hskip\labelsep\scshape #1]\ignorespaces
}{%
	\endtrivlist\@endpefalse
}
\providecommand{\proofname}{Proof}
\fi

\usepackage{hyperref}
\usepackage{url}
\usepackage{amsfonts}
\usepackage{nicefrac}
\usepackage{xcolor}
\usepackage{booktabs}  
\usepackage{diagbox}
\usepackage{subcaption}

\newcommand{\comment}[1]{}

\newcommand{\rowa}[1]{\renewcommand{\arraystretch}{#1}}

\providecommand{\theoremname}{Theorem}

\title{A Unified Wasserstein Distributional Robustness Framework for Adversarial Training}

\author[1]{Tuan Anh Bui}
\author[1]{Trung Le}
\author[2]{Quan Hung Tran}
\author[1]{He Zhao}
\author[1, 3]{Dinh Phung}

\affil[1]{\footnotesize Monash University}
\affil[2]{\footnotesize Adobe Research}
\affil[3]{\footnotesize VinAI Research}

\newcommand{\new}{\marginpar{NEW}}

\iclrfinalcopy 
\begin{document}

\maketitle

\global\long\def\sidenote#1{\marginpar{\small\emph{{\color{Medium}#1}}}}%
\global\long\def\se{\hat{\text{se}}}%
\global\long\def\interior{\text{int}}%
\global\long\def\boundary{\text{bd}}%
\global\long\def\ML{\textsf{ML}}%
\global\long\def\GML{\mathsf{GML}}%
\global\long\def\HMM{\mathsf{HMM}}%
\global\long\def\support{\text{supp}}%
\global\long\def\new{\text{*}}%
\global\long\def\stir{\text{Stirl}}%
\global\long\def\mA{\mathcal{A}}%
\global\long\def\mB{\mathcal{B}}%
\global\long\def\expect{\mathbb{E}}%
\global\long\def\mF{\mathcal{F}}%
\global\long\def\mK{\mathcal{K}}%
\global\long\def\mH{\mathcal{H}}%
\global\long\def\mX{\mathcal{X}}%
\global\long\def\mZ{\mathcal{Z}}%
\global\long\def\mS{\mathcal{S}}%
\global\long\def\Ical{\mathcal{I}}%
\global\long\def\mT{\mathcal{T}}%
\global\long\def\Pcal{\mathcal{P}}%
\global\long\def\dist{d}%
\global\long\def\HX{\entro\left(X\right)}%
\global\long\def\entropyX{\HX}%
\global\long\def\HY{\entro\left(Y\right)}%
\global\long\def\entropyY{\HY}%
\global\long\def\HXY{\entro\left(X,Y\right)}%
\global\long\def\entropyXY{\HXY}%
\global\long\def\mutualXY{\mutual\left(X;Y\right)}%
\global\long\def\mutinfoXY{\mutualXY}%
\global\long\def\given{\mid}%
\global\long\def\gv{\given}%
\global\long\def\goto{\rightarrow}%
\global\long\def\asgoto{\stackrel{a.s.}{\longrightarrow}}%
\global\long\def\pgoto{\stackrel{p}{\longrightarrow}}%
\global\long\def\dgoto{\stackrel{d}{\longrightarrow}}%
\global\long\def\lik{\mathcal{L}}%
\global\long\def\logll{\mathit{l}}%
\global\long\def\bigcdot{\raisebox{-0.5ex}{\scalebox{1.5}{\ensuremath{\cdot}}}}%
\global\long\def\sig{\textrm{sig}}%
\global\long\def\likelihood{\mathcal{L}}%
\global\long\def\vectorize#1{\mathbf{#1}}%
\global\long\def\vt#1{\mathbf{#1}}%
\global\long\def\gvt#1{\boldsymbol{#1}}%
\global\long\def\idp{\ \bot\negthickspace\negthickspace\bot\ }%
\global\long\def\cdp{\idp}%
\global\long\def\das{}%
\global\long\def\id{\mathbb{I}}%
\global\long\def\idarg#1#2{\id\left\{  #1,#2\right\}  }%
\global\long\def\iid{\stackrel{\text{iid}}{\sim}}%
\global\long\def\bzero{\vt 0}%
\global\long\def\bone{\mathbf{1}}%
\global\long\def\a{\mathrm{a}}%
\global\long\def\ba{\mathbf{a}}%
\global\long\def\b{\mathrm{b}}%
\global\long\def\bb{\mathbf{b}}%
\global\long\def\B{\mathrm{B}}%
\global\long\def\boldm{\boldsymbol{m}}%
\global\long\def\c{\mathrm{c}}%
\global\long\def\C{\mathrm{C}}%
\global\long\def\d{\mathrm{d}}%
\global\long\def\D{\mathrm{D}}%
\global\long\def\N{\mathrm{N}}%
\global\long\def\h{\mathrm{h}}%
\global\long\def\H{\mathrm{H}}%
\global\long\def\bH{\mathbf{H}}%
\global\long\def\K{\mathrm{K}}%
\global\long\def\M{\mathrm{M}}%
\global\long\def\bff{\vt f}%
\global\long\def\bx{\mathbf{\mathbf{x}}}%
\global\long\def\bl{\boldsymbol{l}}%
\global\long\def\s{\mathrm{s}}%
\global\long\def\T{\mathrm{T}}%
\global\long\def\bu{\mathbf{u}}%
\global\long\def\v{\mathrm{v}}%
\global\long\def\bv{\mathbf{v}}%
\global\long\def\bo{\boldsymbol{o}}%
\global\long\def\bh{\mathbf{h}}%
\global\long\def\bs{\boldsymbol{s}}%
\global\long\def\x{\mathrm{x}}%
\global\long\def\bx{\mathbf{x}}%
\global\long\def\bz{\mathbf{z}}%
\global\long\def\hbz{\hat{\bz}}%
\global\long\def\z{\mathrm{z}}%
\global\long\def\y{\mathrm{y}}%
\global\long\def\bxnew{\boldsymbol{y}}%
\global\long\def\bX{\boldsymbol{X}}%
\global\long\def\tbx{\tilde{\bx}}%
\global\long\def\by{\mathbf{y}}%
\global\long\def\bY{\boldsymbol{Y}}%
\global\long\def\bZ{\boldsymbol{Z}}%
\global\long\def\bU{\boldsymbol{U}}%
\global\long\def\bn{\boldsymbol{n}}%
\global\long\def\bV{\boldsymbol{V}}%
\global\long\def\bI{\boldsymbol{I}}%
\global\long\def\J{\mathrm{J}}%
\global\long\def\bJ{\mathbf{J}}%
\global\long\def\w{\mathrm{w}}%
\global\long\def\bw{\vt w}%
\global\long\def\bW{\mathbf{W}}%
\global\long\def\balpha{\gvt{\alpha}}%
\global\long\def\bdelta{\boldsymbol{\delta}}%
\global\long\def\bsigma{\gvt{\sigma}}%
\global\long\def\bbeta{\gvt{\beta}}%
\global\long\def\bmu{\gvt{\mu}}%
\global\long\def\btheta{\boldsymbol{\theta}}%
\global\long\def\blambda{\boldsymbol{\lambda}}%
\global\long\def\bgamma{\boldsymbol{\gamma}}%
\global\long\def\bpsi{\boldsymbol{\psi}}%
\global\long\def\bphi{\boldsymbol{\phi}}%
\global\long\def\bpi{\boldsymbol{\pi}}%
\global\long\def\bomega{\boldsymbol{\omega}}%
\global\long\def\bepsilon{\boldsymbol{\epsilon}}%
\global\long\def\btau{\boldsymbol{\tau}}%
\global\long\def\bxi{\boldsymbol{\xi}}%
\global\long\def\realset{\mathbb{R}}%
\global\long\def\realn{\realset^{n}}%
\global\long\def\integerset{\mathbb{Z}}%
\global\long\def\natset{\integerset}%
\global\long\def\integer{\integerset}%
\global\long\def\natn{\natset^{n}}%
\global\long\def\rational{\mathbb{Q}}%
\global\long\def\rationaln{\rational^{n}}%
\global\long\def\complexset{\mathbb{C}}%
\global\long\def\comp{\complexset}%
\global\long\def\compl#1{#1^{\text{c}}}%
\global\long\def\and{\cap}%
\global\long\def\compn{\comp^{n}}%
\global\long\def\comb#1#2{\left({#1\atop #2}\right) }%
\global\long\def\param{\vt w}%
\global\long\def\Param{\Theta}%
\global\long\def\meanparam{\gvt{\mu}}%
\global\long\def\Meanparam{\mathcal{M}}%
\global\long\def\meanmap{\mathbf{m}}%
\global\long\def\logpart{A}%
\global\long\def\simplex{\Delta}%
\global\long\def\simplexn{\simplex^{n}}%
\global\long\def\dirproc{\text{DP}}%
\global\long\def\ggproc{\text{GG}}%
\global\long\def\DP{\text{DP}}%
\global\long\def\ndp{\text{nDP}}%
\global\long\def\hdp{\text{HDP}}%
\global\long\def\gempdf{\text{GEM}}%
\global\long\def\rfs{\text{RFS}}%
\global\long\def\bernrfs{\text{BernoulliRFS}}%
\global\long\def\poissrfs{\text{PoissonRFS}}%
\global\long\def\grad{\gradient}%
\global\long\def\gradient{\nabla}%
\global\long\def\partdev#1#2{\partialdev{#1}{#2}}%
\global\long\def\partialdev#1#2{\frac{\partial#1}{\partial#2}}%
\global\long\def\partddev#1#2{\partialdevdev{#1}{#2}}%
\global\long\def\partialdevdev#1#2{\frac{\partial^{2}#1}{\partial#2\partial#2^{\top}}}%
\global\long\def\closure{\text{cl}}%
\global\long\def\cpr#1#2{\Pr\left(#1\ |\ #2\right)}%
\global\long\def\var{\text{Var}}%
\global\long\def\Var#1{\text{Var}\left[#1\right]}%
\global\long\def\cov{\text{Cov}}%
\global\long\def\Cov#1{\cov\left[ #1 \right]}%
\global\long\def\COV#1#2{\underset{#2}{\cov}\left[ #1 \right]}%
\global\long\def\corr{\text{Corr}}%
\global\long\def\sst{\text{T}}%
\global\long\def\SST{\sst}%
\global\long\def\ess{\mathbb{E}}%
\global\long\def\Ess#1{\ess\left[#1\right]}%
\newcommandx\ESS[2][usedefault, addprefix=\global, 1=]{\underset{#2}{\ess}\left[#1\right]}%
\global\long\def\fisher{\mathcal{F}}%
\global\long\def\bfield{\mathcal{B}}%
\global\long\def\borel{\mathcal{B}}%
\global\long\def\bernpdf{\text{Bernoulli}}%
\global\long\def\betapdf{\text{Beta}}%
\global\long\def\dirpdf{\text{Dir}}%
\global\long\def\gammapdf{\text{Gamma}}%
\global\long\def\gaussden#1#2{\text{Normal}\left(#1, #2 \right) }%
\global\long\def\gauss{\mathbf{N}}%
\global\long\def\gausspdf#1#2#3{\text{Normal}\left( #1 \lcabra{#2, #3}\right) }%
\global\long\def\multpdf{\text{Mult}}%
\global\long\def\poiss{\text{Pois}}%
\global\long\def\poissonpdf{\text{Poisson}}%
\global\long\def\pgpdf{\text{PG}}%
\global\long\def\wshpdf{\text{Wish}}%
\global\long\def\iwshpdf{\text{InvWish}}%
\global\long\def\nwpdf{\text{NW}}%
\global\long\def\niwpdf{\text{NIW}}%
\global\long\def\studentpdf{\text{Student}}%
\global\long\def\unipdf{\text{Uni}}%
\global\long\def\transp#1{\transpose{#1}}%
\global\long\def\transpose#1{#1^{\mathsf{T}}}%
\global\long\def\mgt{\succ}%
\global\long\def\mge{\succeq}%
\global\long\def\idenmat{\mathbf{I}}%
\global\long\def\trace{\mathrm{tr}}%
\global\long\def\argmax#1{\underset{_{#1}}{\text{argmax}} }%
\global\long\def\argmin#1{\underset{_{#1}}{\text{argmin}\ } }%
\global\long\def\diag{\text{diag}}%
\global\long\def\norm{}%
\global\long\def\spn{\text{span}}%
\global\long\def\vtspace{\mathcal{V}}%
\global\long\def\field{\mathcal{F}}%
\global\long\def\ffield{\mathcal{F}}%
\global\long\def\inner#1#2{\left\langle #1,#2\right\rangle }%
\global\long\def\iprod#1#2{\inner{#1}{#2}}%
\global\long\def\dprod#1#2{#1 \cdot#2}%
\global\long\def\norm#1{\left\Vert #1\right\Vert }%
\global\long\def\entro{\mathbb{H}}%
\global\long\def\entropy{\mathbb{H}}%
\global\long\def\Entro#1{\entro\left[#1\right]}%
\global\long\def\Entropy#1{\Entro{#1}}%
\global\long\def\mutinfo{\mathbb{I}}%
\global\long\def\relH{\mathit{D}}%
\global\long\def\reldiv#1#2{\relH\left(#1||#2\right)}%
\global\long\def\KL{KL}%
\global\long\def\KLdiv#1#2{\KL\left(#1\parallel#2\right)}%
\global\long\def\KLdivergence#1#2{\KL\left(#1\ \parallel\ #2\right)}%
\global\long\def\crossH{\mathcal{C}}%
\global\long\def\crossentropy{\mathcal{C}}%
\global\long\def\crossHxy#1#2{\crossentropy\left(#1\parallel#2\right)}%
\global\long\def\breg{\text{BD}}%
\global\long\def\lcabra#1{\left|#1\right.}%
\global\long\def\lbra#1{\lcabra{#1}}%
\global\long\def\rcabra#1{\left.#1\right|}%
\global\long\def\rbra#1{\rcabra{#1}}%

\begin{abstract}
It is well-known that deep neural networks (DNNs) are susceptible to adversarial attacks, 
exposing a severe fragility of deep learning systems. As the result, adversarial training 
(AT) method, by incorporating adversarial examples during training, represents a natural 
and effective approach to strengthen the robustness of a DNN-based classifier. However, 
most AT-based methods, notably PGD-AT and TRADES, typically seek a pointwise adversary 
that generates the worst-case adversarial example by independently perturbing each data 
sample, as a way to ``probe'' the vulnerability of the classifier. Arguably, there are 
unexplored benefits in considering such adversarial effects from an entire distribution. 
To this end, this paper presents a unified framework that connects Wasserstein distributional 
robustness with current state-of-the-art AT methods. We introduce a new Wasserstein cost 
function and a new series of risk functions, with which we show that standard AT methods 
are special cases of their counterparts in our framework. This connection leads to an 
intuitive relaxation and generalization of existing AT methods and facilitates the 
development of a new family of distributional robustness AT-based algorithms. Extensive 
experiments show that our distributional robustness AT algorithms robustify 
further their standard AT counterparts in various settings.\footnote{Our code is available at https://github.com/tuananhbui89/Unified-Distributional-Robustness}
\end{abstract}

\section{Introduction}

Despite remarkable performances of DNN-based deep learning methods,
even the state-of-the-art (SOTA) models are reported to be vulnerable
to adversarial attacks \citep{biggio2013evasion,szegedy2013intriguing,goodfellow2014explaining,madry2017towards,athalye2018obfuscated,zhao2019perturbations,zhao2021learning},
which is of significant concern given the large number of applications
of deep learning in real-world scenarios. Usually, adversarial attacks
are generated by adding small perturbations to benign data but to
change the predictions of the target model. To enhance the robustness
of DNNs, various adversarial defense methods have been developed,
recently \citet{pang2019improving,deng2020adversarial,zhang2020geometry,bai2020improving}.
Among a number of adversarial defenses, Adversarial Training (AT)
is one of the most effective and widely-used approaches \citep{goodfellow2014explaining,madry2017towards,shafahi2019adversarial,tramer2019adversarial,zhang2019defense,xie2020smooth}.
In general, given a classifier, AT can be viewed as a robust optimization
process \citep{ben2009robust} of seeking a pointwise adversary \citep{staib2017distributionally}
that generates the worst-case adversarial example by independently
perturbing each data sample. 

Different from AT, Distributional Robustness (DR) \citep{delage2010distributionally,duchi2021statistics,gao2017wasserstein,gao2016distributionally,rahimian2019distributionally}
looks for a worst-case distribution that generates adversarial examples
from a known uncertainty set of distributions located in the ball
centered around the data distribution. To measure the distance between
distributions, different kinds of metrics have been considered in
DR, such as $f$-divergence \citep{ben2013robust,miyato2015distributional,namkoong2016stochastic}
and Wasserstein distance \citep{shafieezadeh2015distributionally,blanchet2019robust,kuhn2019wasserstein},
where the latter has shown advantages over others on efficiency and
simplicity \citep{staib2017distributionally,sinha2017certifying}.
Therefore, adversary in DR does not look for the perturbation of a
specific data sample, but moves the entire distribution around the
data distribution, thus, is expected to have better generalization
than AT on unseen data samples \citep{staib2017distributionally,sinha2017certifying}.
Conceptually and theoretically, DR can be viewed as a generalization
and better alternative to AT and several attempts \citep{staib2017distributionally,sinha2017certifying}
have shed light on connecting AT with DR. However, to the best of
our knowledge, practical DR approaches that achieve comparable peformance
with SOTA AT methods on adversarial robustness have not been developed
yet.

To bridge this gap, we propose a unified framework that connects distributional
robustness with various SOTA AT methods. Built on
top of Wasserstein Distributional Robustness (WDR), we introduce a
new cost function of the Wasserstein distances and propose a unified
formulation of the risk function in WDR, with which, we can generalize
and encompass SOTA AT methods in the DR setting, including PGD-AT
\citep{madry2017towards}, TRADES \citep{trades}, MART \citep{wang2019improving} and 
AWP \citep{wu2020adversarial}.
With better generalization capacity of distributional robustness,
the resulted AT methods in our DR framework are shown to be able to
achieve better adversarial robustness than their standard
AT counterparts. 

The contributions of this paper are in both theoretical and practical
aspects, summarized as follows: \textbf{1)} Theoretically, we propose
a general framework that bridges distributional robustness and standard
robustness achieved by AT. The proposed framework encompasses the
DR versions of the SOTA AT methods and we prove that these AT methods
are special cases of their DR counterparts. \textbf{2)} Practically,
motivated by our theoretical study, we develop a novel family of algorithms
that generalize the AT methods in the standard robustness setting,
which have better generalization capacity. \textbf{3)} Empirically,
we conduct extensive experiments on benchmark datasets, which show
that the proposed AT methods in the distributional robustness setting
achieve better performance than standard AT methods.

\section{Preliminaries}
\label{sec:pre}

\subsection{Distributional Robustness}

Distributional Robustness (DR) is an emerging framework for learning
and decision-making under uncertainty, which seeks the worst-case
expected loss among a ball of distributions, containing all distributions
that are close to the empirical distribution \citep{gao2017wasserstein}.
As the Wasserstein distance is a powerful and convenient tool of measuring
closeness between distributions, Wasserstein DR has been one of the
most widely-used variant of DR, which has rich applications in (semi)-supervised
learning \citep{blanchet2017semi,chen2018robust,yang2020wasserstein}, generative modeling \citep{ijcai2021-607, dam2019three}, 
transfer learning and domain adaptation \citep{lee2017minimax,duchi2019distributionally,zhao2019learning,nguyen2021tidot, nguyen2021most,le21a_lamda, le2021label}, topic modeling \citep{zhao2021neural}, 
and reinforcement learning \citep{abdullah2019wasserstein,smirnova2019distributionally,derman2020distributional}.
For more comprehensive review, please refer to the surveys of \citet{kuhn2019wasserstein,rahimian2019distributionally}.
Here we consider a generic Polish space $S$ endowed with a distribution
$\mathbb{P}$. Let $f:S\goto\mathbb{R}$ be a real-valued (risk) function
and $c:S\times S\goto\mathbb{R}_{+}$ be a cost function. Distributional
robustness setting aims to find the distribution $\mathbb{Q}$ in
the vicinity of $\mathbb{P}$ and maximizes the risk in the \emph{$\mathbb{E}$}
form \citep{sinha2017certifying,blanchet2019quantifying}: 
\begin{equation}
\sup_{\mathbb{Q}:\mathcal{W}_{c}\left(\mathbb{P},\mathbb{Q}\right)<\epsilon}\mathbb{E}_{\mathbb{Q}}\left[f\left(z\right)\right],\label{eq:primal_form}
\end{equation}
where $\epsilon>0$ and $\mathcal{W}_{c}$ denotes the optimal transport
(OT) cost, or a Wasserstein distance if $c$ is a metric, defined
as: 
\begin{equation}
\mathcal{W}_{c}\left(\mathbb{P},\mathbb{Q}\right):=\inf_{\gamma\in\Gamma\left(\mathbb{P},\mathbb{Q}\right)}\int cd\gamma,\label{eq:asf}
\end{equation}
where $\Gamma\left(\mathbb{P},\mathbb{Q}\right)$ is the set of couplings
whose marginals are $\mathbb{P}$ and $\mathbb{Q}$. With the assumption
that $f\in L^{1}\left(\mathbb{P}\right)$ is upper semi-continuous
and the cost $c$ is a non-negative lower semi-continuous satisfying
$c(z,z')=0\text{ iff }z=z'$, \citet{sinha2017certifying,blanchet2019quantifying}
show that the \emph{dual} form for Eq. (\ref{eq:primal_form}) is:
\begin{gather}
\inf_{\lambda\geq0}\left\{ \lambda\epsilon+\ESS[\sup_{z'}\left\{ f\left(z'\right)-\lambda c\left(z',z\right)\right\} ]{z\sim\mathbb{P}}\right\} .\label{eq:dual_form}
\end{gather}
\citet{sinha2017certifying} further employs a Lagrangian for Wasserstein-based
uncertainty sets to arrive at a relaxed version with $\lambda\geq0$:
\begin{align}
 & \sup_{\mathbb{Q}}\left\{ \mathbb{E}_{\mathbb{Q}}\left[f\left(z\right)\right]-\lambda\mathcal{W}_{c}\left(\mathbb{P},\mathbb{Q}\right)\right\} =\ESS[\sup_{z'}\left\{ f\left(z'\right)-\lambda c\left(z',z\right)\right\} ]{z\sim\mathbb{P}}.\label{eq:dualform_relax}
\end{align}

\subsection{Adversarial Robustness with Adversarial Training\label{subsec:Adversarial-Training}}

In this paper, we are interested in image classification tasks and
focus on the adversaries that add small perturbations to the pixels
of an image to generate attacks based on gradients, which are the
most popular and effective. FGSM \citep{goodfellow2014explaining}
and PGD \citep{madry2017towards} are the most representative gradient-based
attacks and PGD is the most widely-used one, due to its effectiveness
and simplicity. 
Now we consider a classification problem on the space $S=\mathcal{X}\times\mathcal{Y}$
where $\mathcal{X}$ is the data space, $\mathcal{Y}$ is the label
space. 
We would like to learn a classifier that predicts the label of a datum
well $h_{\theta}:\mathcal{X}\goto\mathcal{Y}$. Learning of the classifier
can be done by minimising its loss: $\ell\left(h_{\theta}\left(x\right),y\right)$,
which can typically be the the cross-entropy loss. In addition to
predicting well on benign data, an adversarial defense aims to make the classifier robust
against adversarial examples. As the most successful approach, adversarial
training is a straightforward method that creates and then incorporates
adversarial examples into the training process. With this general
idea, different AT methods vary in the way of picking which adversarial
examples one should train on. Here we list three widely-used AT methods.

\textbf{PGD-AT} \citep{madry2017towards} seeks the \emph{most violating}
examples to improve model robustness:{\small{}
\begin{equation}
\inf_{\theta}\mathbb{E}_{\mathbb{P}}\left[\beta\sup_{x'\in B_{\epsilon}\left(x\right)}CE\left(h_{\theta}\left(x'\right),y\right)+CE\left(h_{\theta}\left(x\right),y\right)\right],\label{eq:pdg-at}
\end{equation}
}where $B_{\epsilon}\left(x\right)=\left\{ x':c_{\mathcal{X}}\left(x,x'\right)\leq\epsilon\right\} $,
$\beta>0$ is the trade-off parameter and cross-entropy loss CE.

\textbf{TRADES} \citep{trades} seeks the \emph{most divergent} examples
to improve model robustness:{\small{}
\begin{equation}
\inf_{\theta}\mathbb{E}_{\mathbb{P}}\left[\beta\sup_{x'}D_{KL}\left(h_{\theta}\left(x'\right),h_{\theta}\left(x\right)\right)+CE\left(h_{\theta}\left(x\right),y\right)\right],\label{eq:trades}
\end{equation}
}where $x'\in B_{\epsilon}\left(x\right)$ and $D_{KL}$ is the usual
Kullback-Leibler (KL) divergence.

\textbf{MART} \citep{wang2019improving} takes into account prediction confidence:{\small{}
\begin{align}
 & \inf_{\theta}\mathbb{E}_{\mathbb{P}}\biggl[\beta\left(1-\left[h_{\theta}\left(x\right)\right]_{y}\right)\sup_{x'\in B_{\epsilon}\left(x\right)}D_{KL}\left(h_{\theta}\left(x'\right),h_{\theta}\left(x\right)\right)+BCE\left(h_{\theta}\left(x\right),y\right)\biggr],\label{eq:mart-1}
\end{align}
}where $BCE\left(h_{\theta}\left(x\right),y\right)$ is defined as:
$-\log\left(\left[h_{\theta}\left(x\right)\right]_{y}\right)-\log\left(1-\max_{k\neq y}\left[h_{\theta}\left(x\right)\right]_{k}\right)$.

\subsection{Connecting Distributional Robustness to Adversarial Training\label{subsec:AML}}

To bridge distributional and adversarial robustness, \citet{sinha2017certifying}
proposes an AT method, named Wasserstein Risk Minimization (WRM),
which generalizes PGD-AT through the principled lens of distributionally
robust optimization. For smooth loss functions, WRM enjoys convergence
guarantees similar to non-robust approaches while certifying performance
even for the worst-case population loss. Specifically, assume that
$\mathbb{P}$ is a joint distribution that generates a pair $z=\left(x,y\right)$
where $x\in\mathcal{X}$ and $y\in\mathcal{Y}$. The cost function
is defined as: $c\left(z,z'\right)=c_{\mathcal{X}}\left(x,x'\right)+\infty\times\boldsymbol{1}\left\{ y\neq y'\right\} $
where $z'=\left(x',y'\right)$, $c_{\mathcal{X}}:\mathcal{X}\times\mathcal{X}\goto\mathbb{R}_{+}$
is a cost function on $\mathcal{X}$, and $\boldsymbol{1}\left\{ \cdot\right\} $
is the indicator function. One can define the risk function $f$ as
the loss of the classifier, i.e., $f\left(z\right):=\ell\left(h_{\theta}\left(x\right),y\right)$.
Together with Eq. (\ref{eq:primal_form}), attaining a robust classifier
is to solve the following min-max problem: 
\begin{equation}
\inf_{\theta}\sup_{\mathbb{Q}:\mathcal{W}_{c}\left(\mathbb{P},\mathbb{Q}\right)<\epsilon}\mathbb{E}_{\mathbb{Q}}\left[\ell\left(h_{\theta}\left(x\right),y\right)\right].\label{eq:primal_AML}
\end{equation}
The above equation shows the generalisation of WRM to PGD-AT. With
Eq. (\ref{eq:dual_form}) and Eq. (\ref{eq:dualform_relax}), one can
arrive at Eq. (\ref{eq:dual_AML_relax}) as below where $\lambda\geq0$ is a trade-off parameter: 
\begin{equation}
\inf_{\theta}\mathbb{E}_{\mathbb{P}}\left[\sup_{x'}\left\{ \ell\left(h_{\theta}\left(x'\right),y\right)- \lambda c_{\mathcal{X}}\left(x',x\right)\right\} \right].\label{eq:dual_AML_relax}
\end{equation}


\section{Proposed Unified Distribution Robustness Framework\label{sec:Proposed-Unified-Distribution}}

Although WRM~\citep{sinha2017certifying} sheds light on connecting
distributional robustness with adversarial training, its framework
and formulation is limited to PGD-AT, which cannot encompass more
advanced AT methods including TRADES and MART. In this paper, we propose
a unified formulation for distributional robustness, which is a more
general framework connecting state-of-the-art AT and existing distributional
robustness approaches where they become special cases.

Let $\mathbb{P}^{d}$ be the data distribution that generates instance
$x\sim\mathbb{P}^{d}$ and $\mathbb{P}_{.|x}^{l}$ the conditional
to generate label $y\sim\mathbb{P}_{.|x}^{l}$ given $x$ where $x\in\mathcal{X},y\in\mathcal{Y}$.
For our purpose, we consider the space $S=\mathcal{X}\times\mathcal{X}\times\mathcal{Y}$
and a joint distribution $\mathbb{P}_{\triangle}$ on $S$ consisting
of samples $\left(x,x,y\right)$ where $x\sim\mathbb{P}^{d}$ and
$y\sim\mathbb{P}_{.|x}^{l}$. Now consider a distribution $\mathbb{Q}$
on $S$ such that $\mathcal{W}_{c}\left(\mathbb{Q},\mathbb{P}_{\triangle}\right)<\epsilon$.
A draw $z\sim\mathbb{P}_{\triangle}$ will take the form $z=\left(x,x,y\right)$
whereas $z'\sim\mathbb{Q}$ will be $z'=\left(x',x'',y'\right)$.
We propose cost function $c(z,z')$ defined as: 
\begin{equation}
c(z,z')=c_{\mathcal{X}}\left(x,x'\right)+\infty\times c_{\mathcal{X}}\left(x,x''\right)+\infty\times\boldsymbol{1}\left\{ y\neq y'\right\} ,\label{eq:udr_cost}
\end{equation}
where we note that this cost function is non negative, satisfies $c(z,z)=0$ and lower semi-continuous, i.e., $ \underset{z' \rightarrow z_0}{\lim} \inf c(z,z') \geq c(z,z_0) $.

With our new setting, it is useful to understand the ``vicinity''of
\textbf{$\mathbb{P}_{\Delta}$} via the distribution OT-ball condition
$\mathcal{W}_{c}\left(\mathbb{Q},\mathbb{P}_{\triangle}\right)<\epsilon$.
Since there exists a transport plan $\gamma\in\Gamma\left(\mathbb{P}_{\triangle},\mathbb{Q}\right)$
s.t. $\int cd\gamma<\epsilon$ and $c(z,z')$ is finite a.s. $\gamma$,
this implies that if $\left(z,z'\right)\sim\gamma$, then first, it
is easy to see that $x''=x$ and $y'=y$, and second, $x'$ tends
to be close to $x$. To see why the later is the case, since $\mathbb{P}^{d}$
is a marginal of $\mathbb{P}_{\triangle}$ on the first $x$ in $\left(x,x,y\right)$,
therefore if $\mathbb{Q}^{d}$ is the marginal of $\mathbb{Q}$ on
$x'$ in $(x',x'',y')$ then $\mathcal{W}_{d}\left(\mathbb{Q}^{d},\mathbb{P}^{d}\right)\leq\mathcal{W}_{d}\left(\mathbb{Q},\mathbb{P}_{\triangle}\right)<\epsilon$,
which explains the closeness between of $x$ and $x'$.

Given $z'=\left(x',x'',y'\right)\sim\mathbb{Q}$ where $\mathcal{W}_{c}\left(\mathbb{Q},\mathbb{P}_{\triangle}\right)<\epsilon$,
we define a unified risk function $g_{\theta}\left(z'\right)$ w.r.t
a classifier $h_{\theta}$ that encompasses the unified distributional
robustness (UDR) version for PGD-AT, TRADES, and MART (cf Section
\ref{subsec:Adversarial-Training}): 
\begin{itemize}
\item \emph{UDR-PGD}: $g_{\theta}\left(z'\right):=CE\left(h_{\theta}\left(x''\right),y'\right)+\beta CE\left(h_{\theta}\left(x'\right),y'\right)$. 
\item \emph{UDR-TRADES}: $g_{\theta}\left(z'\right):=CE\left(h_{\theta}\left(x''\right),y'\right)+\beta D_{KL}\left(h_{\theta}\left(x'\right),h_{\theta}\left(x''\right)\right)$. 
\item \emph{UDR-MART}: $g_{\theta}\left(z'\right):=BCE\left(h_{\theta}\left(x''\right),y'\right)+\beta(1-\left[h_{\theta}\left(x''\right)\right]_{y})D_{KL}\left(h_{\theta}\left(x'\right),h_{\theta}\left(x''\right)\right).$\footnote{To encompass MART with our framework, we assume a classifier is adversarially trained by Eq.~(\ref{eq:mart-1}) with adversarial examples generated by $\sup_{x'\in B_{\epsilon}\left(x\right)}D_{KL}\left(h_{\theta}\left(x'\right),h_{\theta}\left(x\right)\right)+BCE\left(h_{\theta}\left(x\right),y\right)$. This is slightly different from the original MART, where the adversarial examples are generated by $\sup_{x'\in B_{\epsilon}\left(x\right)} CE\left(h_{\theta}\left(x'\right),y\right)$.} 
\end{itemize}

Now we derive the primal and dual objectives for the proposed UDR
framework. With the UDR risk function $g_{\theta}(z')$ defined previously,
following Eq. (\ref{eq:primal_form}) and Eq. (\ref{eq:dual_form}),
the primal (left) and dual (right) forms of our UDR objective are:
\begin{equation}
\inf_{\theta}\sup_{\mathbb{Q}:\mathcal{W}_{c}\left(\mathbb{Q},\mathbb{P}_{\triangle}\right)<\epsilon}\mathbb{E}_{\mathbb{Q}}\left[g_{\theta}\left(z'\right)\right]=\inf_{\theta}\inf_{\lambda\geq0}\left(\lambda\epsilon+\mathbb{E}_{\mathbb{P}_{\triangle}}\left[\sup_{z'}\left\{ g_{\theta}\left(z'\right)-\lambda c\left(z',z\right)\right\} \right]\right).\label{eq:udr_primal_dual}
\end{equation}
With the cost function $c$ defined in Eq. (\ref{eq:udr_cost}), the
dual form in (\ref{eq:udr_primal_dual}) can be rewritten as: 
\begin{gather}
\inf_{\theta,\lambda\geq0}\left(\lambda\epsilon+\mathbb{E}_{\mathbb{P}_{\simplex}}\left[\sup_{x',x''=x,y'=y}\left\{ g_{\theta}\left(z'\right)-\lambda c_{\mathcal{X}}\left(x',x\right)\right\} \right]\right)=\nonumber \\
\inf_{\theta,\lambda\geq0}\left(\lambda\epsilon+\mathbb{E}_{\mathbb{P}}\left[\sup_{x'}\left\{ g_{\theta}\left(x',x,y\right)-\lambda c_{\mathcal{X}}\left(x',x\right)\right\} \right]\right)\label{eq:udr_simple_dual}
\end{gather}
where we note that $\mathbb{P}$ is a distribution over pairs $\left(x,y\right)$
for which $x\sim\mathbb{P}^{d}$ and $y\sim\mathbb{P}_{.|x}^{l}$.
The min-max problem in Eq. (\ref{eq:udr_simple_dual}) encompasses
the PGD-AT, TRADES, and MART distributional robustness counterparts on the choice of the function $g_{\theta}\left(x',x,y\right)$
by simply choosing an appropriate $g_{\theta}\left(x',x,y\right)$
as shown in Section \ref{subsec:AML}.


In what follows, we prove that standard PGD-AT, TRADES, and MART presented
in Section \ref{sec:pre} are specific cases of their UDR counterparts
by specifying corresponding cost functions. Given a cost function
$c_{\mathcal{X}}$ (e.g., $L_{1}$, $L_{2}$, and $L_{\infty}$),
we define a new cost function $\tilde{c}_{\mathcal{X}}$ as: 
\begin{equation}
\tilde{c}_{\mathcal{X}}\left(x,x'\right)=\begin{cases}
c_{\mathcal{X}}\left(x,x'\right) & \text{if}\,c_{\mathcal{X}}(x,x')\leq\epsilon\\
\infty & \text{otherwise}.
\end{cases}\label{eq:c_til}
\end{equation}
The cost function $\tilde{c}_{\mathcal{X}}$ is lower semi-continuous.
By defining the ball $B_{\epsilon}\left(x\right):=\left\{ x':c_{\mathcal{X}}\left(x,x'\right)\leq\epsilon\right\} =\left\{ x':\tilde{c}_{\mathcal{X}}\left(x,x'\right)\leq\epsilon\right\} $
, we achieve the following theorem on the relation between distributional
and standard robustness. 
\begin{thm}
\label{thm:relation}With the cost function $\tilde{c}_{x}$ defined
as above, the optimization problem: 
\begin{equation}
\inf_{\theta,\lambda\geq0}\left(\lambda\epsilon+\mathbb{E}_{\mathbb{P}}\left[\sup_{x'}\left\{ g_{\theta}\left(x',x,y\right)-\lambda\tilde{c}_{\mathcal{X}}\left(x',x\right)\right\} \right]\right)\label{eq:udr_unified_theo}
\end{equation}
is equivalent to the optimization problem: 
\begin{equation}
\inf_{\theta}\mathbb{E}_{\mathbb{P}}\left[\sup_{x'\in B_{\epsilon}\left(x\right)}g_{\theta}\left(x',x,y\right)\right].\label{eq:std_unified}
\end{equation}
\end{thm}

\begin{proof}
See Appendix \ref{sec:Proof} for the proof.
        \vspace{-3mm}
\end{proof}

\textbf{Theoretical contribution and comparison to previous work.~}
Theorem \ref{thm:relation} says that the standard PGD-AT, TRADES,
and MART are special cases of their UDR counterparts, which indicates
that our UDR versions of AT have a richer expressiveness capacity
than the standard ones. Different from WRM \citep{sinha2017certifying}
, our proposed framework is developed based on theoretical foundation
of \citep{blanchet2019quantifying}. It is worth noting that the theoretical
development is\emph{ not trivial} because theory developed in \citet{blanchet2019quantifying}
is only valid for a bounded cost function, while the cost function
$\tilde{c}$ is unbounded. More specifically, the transformation from
primal to dual forms in Eq. (\ref{eq:udr_primal_dual}) requires the
cost function $c$ to be bounded. In Theorem \ref{thm:primal_dual_form}
in Appendix \ref{sec:Proof}, we prove this primal-dual form transformation
for the unbounded cost function $\tilde{c}_{\mathcal{X}}$, which
is certainly not trivial.


Moreover, our UDR is \emph{fundamentally distinctive} from WRM 
in its ability to adapt and learn $\lambda$, while this is a hyper-parameter
in WRM. As a result of a fixed $\lambda$, WRM is fundamentally same
as PGD in the sense that these methods can only utilize
local information of relevant benign examples when crafting adversarial
examples. In contrast, our UDR can leverage both local and global
information of multiple benign examples when crafting adversarial examples
due to the fact that $\lambda$ is adaptable and captures the global
information when solving the outer minimization in (\ref{eq:udr_unified_theo}).
Further explanation can be found in Appendix \ref{sec:global_information}.

\section{Learning Robust Models with UDR}


In this section we introduce the details of how to learn robust models with UDR.
To do this, we first discuss the 
induced cost function $\tilde{c}_{\mathcal{X}}$ defined as in Eq
(\ref{eq:c_til}), which assists us in understanding the connection
between distributional and standard robustness approaches. We note
that $\tilde{c}_{\mathcal{X}}$ is non-differential outside the perturbation
ball (i.e., $c_{\mathcal{X}}(x',x)\geq\epsilon$). To circumvent this,
we introduce a smoothed version $\hat{c}_{\mathcal{X}}$ to approximate
$\tilde{c}_{\mathcal{X}}$ as follows: 
\begin{gather}
\hat{c}_{\mathcal{X}}\left(x,x'\right):=\boldsymbol{1}\left\{ c_{\mathcal{X}}(x,x')<\epsilon\right\} c_{\mathcal{X}}(x,x')+\boldsymbol{1}\left\{ c_{\mathcal{X}}(x,x')\geq\epsilon\right\} \left(\epsilon+\frac{c_{\mathcal{X}}(x,x')-\epsilon}{\tau}\right),\label{eq:smoothed_cost}
\end{gather}
where $\tau>0$ is the temperature to control the growing rate of
the cost function when $x'$ goes out of the perturbation ball. It
is obvious that $\hat{c}_{\mathcal{X}}\left(x,x'\right)$ is continuous
and approaches $\tilde{c}_{\mathcal{X}}\left(x,x'\right)$ when $\tau\goto0$.
Using the smoothed function $\hat{c}_{\mathcal{X}}\left(x,x'\right)$
from Eq. (\ref{eq:smoothed_cost}), the final object of our UDR becomes:
\begin{equation}
\inf_{\theta,\lambda\geq0}\left(\lambda\epsilon+\mathbb{E}_{\mathbb{P}}\left[\sup_{x'}\left\{ g_{\theta}\left(x',x,y\right)-\lambda\hat{c}_{\mathcal{X}}\left(x',x\right)\right\} \right]\right).\label{eq:final_unified}
\end{equation}
With this final objective, our training strategy involves three iterative
steps at each iteration w.r.t. a batch of data examples, which are
shown in Algorithm~\ref{alg}.

\begin{wrapfigure}{R}{0.6\textwidth}
    \begin{minipage}{0.6\textwidth}
        \begin{algorithm}[H]
            \textbf{Input}: training set $\mathcal{D}$, number of iterations $T$, batch size $N$, adversary parameters $\{k,\epsilon,\eta\}$
            
            \textbf{for }$t=1$ to $T$ \textbf{do} 
            \begin{enumerate}
            \item \textbf{Sample} mini-batch $\{x_{i},y_{i}\}_{i=1}^{N}\sim\mathcal{D}$ 
            \item \textbf{Find} adversarial examples $\{x_{i}^{a}\}_{i=1}^{N}$ using
            Eq. (\ref{eq:find-xa}) 
            \begin{enumerate}
            \item \textbf{Initialize} randomly: $x_{i}^{0}=x_{i}+noise$ where $noise\sim\mathcal{U}(-\epsilon,\epsilon)$ 
            \item \textbf{for} $n=1$ to $k$ \textbf{do} 
            \begin{enumerate}
                \item $x_{i}^{inter}=x_{i}^{n}+\eta\text{sign}\left(\nabla_{x}g_{\theta}(x_{i}^{n},x_{i},y_{i})\right)$ 
                \item $x_{i}^{n+1}=x_{i}^{inter}-\eta\lambda\nabla_{x}\hat{c}(x_{i}^{inter},x_{i})$ 
            \end{enumerate}
            \item \textbf{Clip} to valid range: $x_{i}^{a}=clip(x_{i}^{k},0,1)$ 
            \end{enumerate}
            \item \textbf{Update} parameter $\lambda$ using Eq. (\ref{eq:update-lamda}) 
            \item \textbf{Update} model parameter $\theta$ using Eq. (\ref{eq:update-theta}) 
            \end{enumerate}
            \textbf{Output: }model parameter\textbf{ $\theta$}
            \caption{The pseudocode of our proposed method.}
            \label{alg}   
        \end{algorithm}
    \end{minipage}
            \vspace{-5mm}
\end{wrapfigure}

\textbf{1. Craft adversarial examples w.r.t. the current model and the parameter
$\lambda$.~}
Given the current model $\theta$ and the parameter $\lambda$, we
find the adversarial examples by solving: 
\begin{equation}
x^{a}=\text{argmax}_{x'}\left\{ g_{\theta}(x',x,y)-\lambda\hat{c}_{\mathcal{X}}\left(x',x\right)\right\} ,\label{eq:find-xa}
\end{equation}
where different methods (i.e., UDR-PGD, UDR-TRADES, etc.) specifies
$g_{\theta}(x',x,y)$ differently.

Similar to other AT methods like PGD-AT, we employ iterative gradient
ascent update steps to optimise to find $x^{a}$. Specifically, we
start from a random example inside the ball $B_{\epsilon}$ and update
in $k$ steps with the step size $\eta>0$. Since the magnitude of
the gradient $\nabla_{x'}g_{\theta}(x',x,y)$ is significantly smaller
than that of $\nabla_{x'}\hat{c}_{\ \mathcal{X}}(x',x)$, we use $\text{sign}\left(\nabla_{x'}\hat{c}_{\mathcal{X}}(x',x)\right)$
in the update formula rather than $\nabla_{x'}\hat{c}_{\mathcal{X}}(x',x)$.
These steps are shown in 2(a) to 2(c) of Algorithm~\ref{alg}.

An important difference from ours to other AT methods is that at each
update step, we do not apply any explicit projecting operations onto
the ball $B_{\epsilon}$. Indeed, the parameter $\lambda$ controls
how distant $x^{a}$ to its benign counterpart $x$. Thus, this can
be viewed as implicitly projecting onto a soft ball governed by the
magnitude of the parameter $\lambda$ and the temperature $\tau$.
Specifically, when $\lambda$ becomes higher, the crafted adversarial
examples $x^{a}$ stay closer to their benign counterparts $x$ and
vice versa. When $\tau$ is set closer to $0$, the smoothed
cost function $\hat{c}_{\mathcal{X}}$ approximates the cost function
$\tilde{c}_{\mathcal{X}}$ more tightly. Thus, our soft-ball projection
is more identical to the hard ball projection as in projected gradient
ascent.


\looseness=-1 

\textbf{2. Update the parameter $\lambda$.~}
Given current model $\theta$, we craft a batch of adversarial examples
$\{x_{i}^{a}\}_{i=1}^{N}$ corresponding to the benign examples $\left\{ x_{i}\right\} _{i=1}^{N}$
crafted as above. Inspired by the Danskin's theorem , we update $\lambda$
as follows: 
\begin{equation}
\lambda_{n}=\lambda-\eta_{\lambda}\left(\epsilon-\frac{1}{N}\sum_{i=1}^{N}\hat{c}_{\mathcal{X}}(x_{i}^{a},x_{i})\right),\label{eq:update-lamda}
\end{equation}
where $\eta_{\lambda}>0$ is a learning rate and $\lambda_{n}$ represents
the new value of $\lambda$.

The proposed update of $\lambda$ is intuitive: 
\emph{if the adversarial examples stay close to their benign examples,
i.e., $\sum_{i=1}^{N}\hat{c}_{\mathcal{X}}(x_{i}^{a},x_{i})<\epsilon$,
$\lambda$ decreases to make them more distant to the benign examples
and vice versa}. Therefore the adversarial examples are crafted more
diversely, which can further strengthen the robustness of the model.

\textbf{3. Update the model parameter $\theta$.~}
Given the set of adversarial examples $\{x_{i}^{a}\}_{i=1}^{N}$ crafted
as above and their benign examples $\left\{ x_{i}\right\} _{i=1}^{N}$
with the labels $\left\{ y_{i}\right\} _{i=1}^{N}$, we update the
model parameter $\theta$ to minimize $\mathbb{E}_{\mathbb{P}}\left[\nabla g_{\theta}(x^{a},x,y)\right]$
using the current batches of adversarial and benign examples: 
\begin{equation}
\theta_{n}=\theta-\frac{\eta_{\theta}}{N}\sum_{i=1}^{N}\nabla_{\theta}g_{\theta}(x_{i}^{a},x_{i},y_{i}),\label{eq:update-theta}
\end{equation}
where $\eta_{\theta}>0$ is a learning rate and $\theta_{n}$ specifies
the new model parameter.

\looseness=-1

\section{Experiments}



We use MNIST \citep{mnist}, CIFAR10 and CIFAR100 \citep{cifar10}
as the benchmark datasets in our experiment. The inputs were normalized
to $[0,1]$. We apply padding 4 pixels at all borders before random
cropping and random horizontal flips as used in \citet{trades}. We
use both standard CNN architecture \citep{carlini2017towards} and
ResNet architecture \citep{he2016deep} in our experiment. The architecture
and training setting are provided in Appendix \ref{sec:Exp-Setting}.


We compare our UDR with the SOTA AT methods, i.e., \textbf{PGD-AT}
\citep{madry2017towards}, \textbf{TRADES} \citep{trades} and \textbf{MART}
\citep{wang2019improving}. Because TRADES and MART performances are
strongly dependent on the trade-off ratio (i.e., $\beta$ in Eq. (\ref{eq:trades})
and (\ref{eq:mart-1})) between natural loss and robust loss, we use
the original setting in their papers (CIFAR10/CIFAR100: $\beta=6$
for TRADES/UDR-TRADES, $\beta=5$ for MART/UDR-MART; MNIST: $\beta=1$
for all the methods). 
We also tried with the distributional robustness
method WRM \citep{sinha2017certifying}. However, WRM did not seem to obtain reasonable performance in our experiments. 
Its results can be found in Appendix \ref{sec:supp-exp}.
For all the AT methods, we use $\{k=40,\epsilon=0.3,\eta=0.01\}$
for the MNIST dataset, $\{k=10,\epsilon=8/255,\eta=2/255\}$ for the
CIFAR10 dataset and $\{k=10,\epsilon=0.01,\eta=0.001\}$ for the CIFAR100
dataset, where $k$ is number of iteration, $\epsilon$ is the distortion
bound and $\eta$ is the step size of the adversaries.


We use different SOTA attacks to evaluate the defense methods including:
\textbf{1) PGD attack} \citep{madry2017towards} which is one of the
most widely-used gradient based attacks. For PGD, we set $k=200$
and $\epsilon=0.3,\eta=0.01$ for MNIST, $\epsilon=8/255,\eta=2/255$
for CIFAR10, and $\epsilon=0.01,\eta=0.001$ for CIFAR100, which are
the standard settings. \textbf{2) B\&B attack} \citep{brendel2019accurate}
which is a decision based attack. Following \citet{tramer2020adaptive},
we initialized with the PGD attack with $k=20$ and corresponding
$\{\epsilon,\eta\}$ then apply B\&B attack with 200 steps. \textbf{3)
Auto-Attack (AA)} \citep{croce2020reliable} which is an ensemble
methods of four different attacks. We use $\epsilon=0.3,8/255,0.01$,
for MNIST, CIFAR10, and CIFAR100, respectively. The distortion metric
we use in our experiments is $l_{\infty}$ for all measures. We use
the full test set for PGD and 1000 test samples for the other attacks.

\subsection{Main Results}
\begin{table}
    \caption{Comparisons of natural classification accuracy (Nat) and adversarial
    accuracies against different attacks. Best scores are highlighted
    in boldface.\label{tab:whitebox}}
    \centering \resizebox{0.9\linewidth}{!}{ %
    \begin{tabular}{@{}ccccccccccccccc@{}}
    \toprule 
     & \multicolumn{4}{c}{MNIST} &  & \multicolumn{4}{c}{CIFAR10} &  & \multicolumn{4}{c}{CIFAR100}\tabularnewline
    \midrule 
     & Nat  & PGD  & AA  & B\&B  &  & Nat  & PGD  & AA  & B\&B  &  & Nat  & PGD  & AA  & B\&B\tabularnewline
    PGD-AT  & 99.4  & 94.0  & 88.9  & 91.3  &  & \textbf{86.4}  & 46.0  & 42.5  & 44.2  &  & 72.4  & 41.7  & 39.3  & 39.6\tabularnewline
    UDR-PGD  & \textbf{99.5}  & \textbf{94.3}  & \textbf{90.0}  & \textbf{91.4}  &  & \textbf{86.4}  & \textbf{48.9}  & \textbf{44.8}  & \textbf{46.0}  &  & \textbf{73.5}  & \textbf{45.1}  & \textbf{41.9}  & \textbf{42.3}\tabularnewline
    \midrule 
    TRADES  & \textbf{99.4}  & 95.1  & 90.9  & 92.2  &  & 80.8  & 51.9  & 49.1  & 50.2  &  & 68.1  & 49.7  & 46.7  & 47.2\tabularnewline
    UDR-TRADES  & \textbf{99.4}  & \textbf{96.9}  & \textbf{92.2}  & \textbf{95.2}  &  & \textbf{84.4}  & \textbf{53.6}  & \textbf{49.9}  & \textbf{51.0}  &  & \textbf{69.6}  & \textbf{49.9}  & \textbf{47.8}  & \textbf{48.7}\tabularnewline
    \midrule 
    MART  & \textbf{99.3}  & 94.7  & 90.6  & 92.9  &  & \textbf{81.9}  & 53.3  & 48.2  & 49.3  &  & \textbf{68.1}  & 49.8  & 44.8  & 45.4\tabularnewline
    UDR-MART  & \textbf{99.3}  & \textbf{96.0}  & \textbf{92.3}  & \textbf{94.4}  &  & 80.1  & \textbf{54.1}  & \textbf{49.1}  & \textbf{50.4}  &  & 67.5  & \textbf{52.0}  & \textbf{48.5}  & \textbf{48.6}\tabularnewline
    \bottomrule
    \end{tabular}}
    \end{table}
\begin{figure}
    \centering %
    \begin{minipage}[c]{0.62\textwidth}%
    \rowa{1.3} \centering \captionof{table}{Robustness evaluation
    under different PGD attack strengths $\epsilon$. \emph{Avg} represents
    for the average improvement of our DR methods over their counterparts.
    } \label{tab:muli-eps} \resizebox{0.85\linewidth}{!}{ %
    \begin{tabular}{@{}cccccccc@{}}
    \toprule 
    \multicolumn{8}{c}{MNIST}\tabularnewline
    $\epsilon$  & 0.3  & 0.325  & 0.35  & 0.375  & 0.4  & 0.425  & Avg \tabularnewline
    \midrule 
    PGD-AT  & 94.0  & 67.8  & 21.1  & 6.8  & 2.3  & 1.2  & -\tabularnewline
    UDR-PGD  & \textbf{94.3}  & \textbf{92.9}  & \textbf{90.1}  & \textbf{79.2}  & \textbf{22.3}  & \textbf{3.8}  & 31.57\tabularnewline
    \midrule 
    TRADES  & 95.5  & 85.2  & 34.4  & 5.8  & 0.6  & 0.1  & -\tabularnewline
    UDR-TRADES  & \textbf{96.9}  & \textbf{96.9}  & \textbf{95.8}  & \textbf{95.1}  & \textbf{94.5}  & \textbf{88.5}  & 57.68\tabularnewline
    \midrule 
    MART  & 94.7  & 66.1  & 9.4  & 0.9  & 0.2  & 0.1  & -\tabularnewline
    UDR-MART  & \textbf{96.0}  & \textbf{95.0}  & \textbf{94.1}  & \textbf{92.8}  & \textbf{88.8}  & \textbf{37.7}  & 55.5\tabularnewline
    \multicolumn{8}{c}{CIFAR10}\tabularnewline
    $\epsilon$  & $\frac{8}{255}$  & $\frac{10}{255}$  & $\frac{12}{255}$  & $\frac{14}{255}$  & $\frac{16}{255}$  & $\frac{20}{255}$  & Avg\tabularnewline
    \midrule 
    PGD-AT  & 46.0  & 33.7  & 23.7  & 15.2  & 9.5  & 3.6  & -\tabularnewline
    UDR-PGD  & \textbf{48.9}  & \textbf{36.4}  & \textbf{26.3}  & \textbf{18.5}  & \textbf{13.0}  & \textbf{7.1}  & 3.08\tabularnewline
    \midrule 
    TRADES  & 51.9  & 42.5  & 33.7  & 25.7  & 18.9  & 9.1  & -\tabularnewline
    UDR-TRADES  & \textbf{53.6 } & \textbf{43.6}  & \textbf{35.2}  & \textbf{27.5}  & \textbf{20.7}  & \textbf{10.9}  & 1.62\tabularnewline
    \midrule 
    MART  & 53.3  & 43.2  & 34.1  & 25.5  & 18.4  & 9.0  & -\tabularnewline
    UDR-MART  & \textbf{54.1}  & \textbf{46.0}  & \textbf{37.3}  & \textbf{29.7}  & \textbf{22.9}  & \textbf{12.2}  & 3.12\tabularnewline
    \multicolumn{8}{c}{CIFAR100}\tabularnewline
    $\epsilon$  & $\frac{10}{1000}$  & $\frac{12.5}{1000}$  & $\frac{15}{1000}$  & $\frac{17.5}{1000}$  & $\frac{20}{1000}$  & $\frac{25}{1000}$  & Avg\tabularnewline
    \midrule 
    PGD-AT  & 41.7  & 34.5  & 27.8  & 22.6  & 18.2  & 11.7  & -\tabularnewline
    UDR-PGD  & \textbf{45.1}  & \textbf{38.3}  & \textbf{31.9}  & \textbf{26.2}  & \textbf{21.4}  & \textbf{14.2}  & 3.43\tabularnewline
    \midrule 
    TRADES  & 49.7  & 44.3  & 39.9  & 35.2  & 31.2  & 23.5  & -\tabularnewline
    UDR-TRADES  & \textbf{49.9}  & \textbf{44.8}  & \textbf{40.3}  & \textbf{35.7}  & \textbf{31.7}  & \textbf{24.2}  & 0.47\tabularnewline
    \midrule 
    MART  & 49.8  & 45.3  & 41.0  & 36.6  & 32.4  & 25.1  & -\tabularnewline
    UDR-MART  & \textbf{52.0}  & \textbf{47.8}  & \textbf{44.1}  & \textbf{40.2}  & \textbf{36.2}  & \textbf{29.4}  & 3.25\tabularnewline
    \bottomrule
    \end{tabular}} 
    \end{minipage}\hspace{0.005\textwidth} %
    \begin{minipage}[c]{0.36\textwidth}%
    \begin{subfigure}[b]{0.9\linewidth} 
        \centering
        \includegraphics[width=0.99\textwidth]{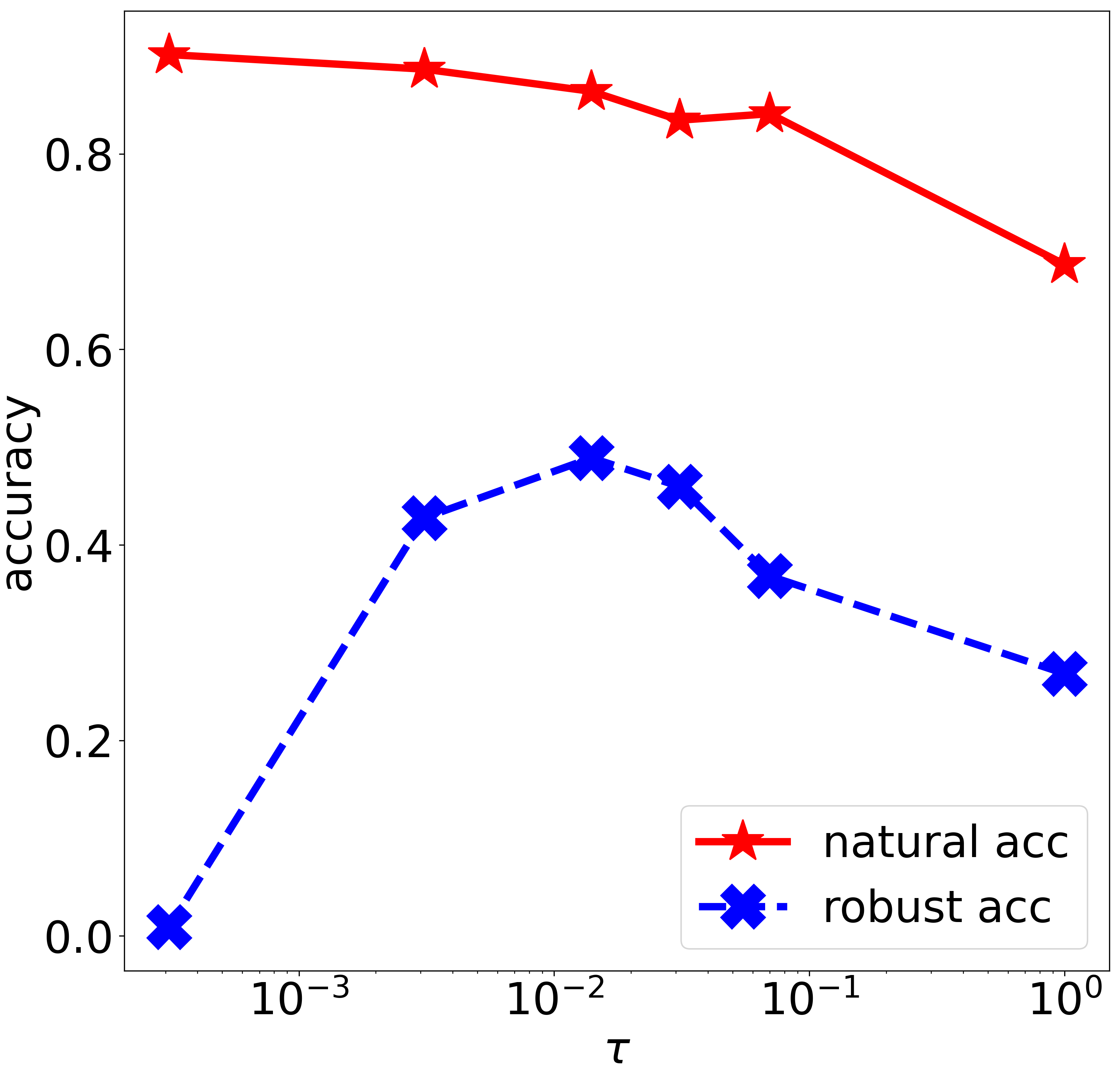}
        \caption{Natural/robust accuracy trade-off\label{fig:tau-tradeoff}}
        \end{subfigure} 
    \begin{subfigure}[b]{0.9\linewidth} 
        \centering
        \includegraphics[width=0.99\textwidth]{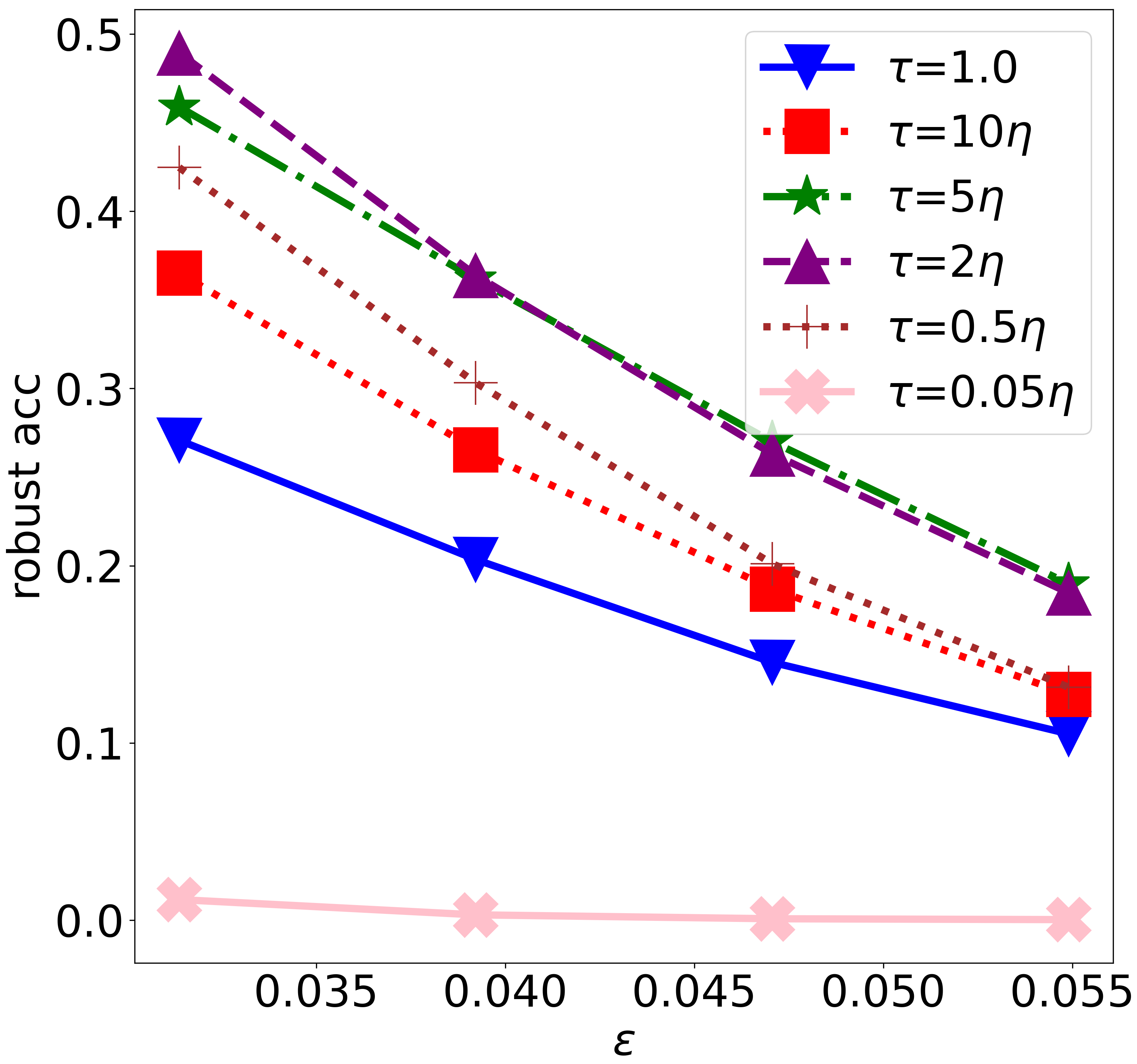}
        \caption{Robustness in correlation with $\tau$ \label{fig:tau-multieps}}
        \end{subfigure} 
    \caption{Further analysis on parameter sensitivity.}
    \label{fig:fa} %
    \end{minipage}
    \vspace{-5mm}
\end{figure}
\textbf{Whitebox Attacks with fixed $\epsilon$.~}
First, we compare the natural and robust accuracy of the AT methods
and their counterparts under our UDR framework, against several SOTA
attacks. Note that in this experiment, the attacks are with their
standard 
settings. The result of this experiment is shown in Table~\ref{tab:whitebox}.
It can be observed that for all the AT methods, our UDR versions are
able to boost the model robustness significantly against all the strong
attack methods in comparison on all the three datasets. These improvements
clearly show that our UDR empowered AT methods achieve the SOTA adversarial
robustness performance. Specifically, our UDR-PGD's improvement over
PGD on both CIFAR10 and CIFAR100 is over 3\% against all the attacks.
Similarly, our UDR-MART also improves over MART with a 3\% gap on
CIFAR100. 

\textbf{Whitebox Attacks with varied $\epsilon$.~}
Recall that UDR is designed to have better generalization capacity
than standard adversarial robustness. In this experiment, we exam
the generalization capacity by attacking the AT methods (including
our UDR variants) with PGD with varied attack strength $\epsilon$
while keeping other parameters of PGD attack the same. This is a highly
practical scenario where attackers may use various attack strengths
that are different from that the model is trained with. The results
of this experiment are shown in Table~\ref{tab:muli-eps}. We have
the following remarks of the results: \textbf{1)} All AT methods perform
reasonably well (our UDR variants are better than their counterparts)
when PGD attacks with the same $\epsilon$ that these methods are
trained on. This is shown in the first column on all the datasets,
whose results are in line with these in Table~\ref{tab:whitebox}.
\textbf{2)} With increased $\epsilon$, the performance of all the
AT methods deteriorates, which is natural. However, the advantage
of our UDR methods over their counterparts becomes more and more significant.
For example, when $\epsilon=0.375$, all of our UDR methods can achieve
at least 80\% robust accuracy on MNIST, while others can barely defend.
This clearly demonstrates the benefit of our UDR framework on generalization
capacity.

\begin{table}
    \caption{Adversarial accuracy in the blackbox settings. \emph{Avg} represents
    for the average improvement of our DR methods over their counterparts.\label{tab:blackbox}}
    \centering \resizebox{0.75\linewidth}{!}{ %
    \begin{tabular}{@{}cccccccc@{}}
    \toprule 
    \backslashbox{Target}{Source}  & PGD-AT  & UDR-P  & TRADES  & UDR-T  & MART  & UDR-M  & Avg\tabularnewline
    \midrule 
    PGD-AT  & -  & -  & 61.6  & 61.6  & 61.7  & 62.4  & -\tabularnewline
    UDR-PGD  & -  & -  & \textbf{63.6}  & \textbf{63.4}  & \textbf{64.0}  & \textbf{64.1}  & 2.0\tabularnewline
    \midrule 
    TRADES  & 61.2  & 61.3  & -  & -  & 58.9  & 59.8  & -\tabularnewline
    UDR-TRADES  & \textbf{62.7}  & \textbf{62.8}  & -  & -  & \textbf{61.1}  & \textbf{61.6}  & 1.8\tabularnewline
    \midrule 
    MART  & 61.4  & 61.4  & 58.9  & 59.5  & -  & -  & -\tabularnewline
    UDR-MART  & \textbf{62.3}  & \textbf{62.1}  & \textbf{60.1}  & \textbf{60.5}  & -  & -  & 1.0\tabularnewline
    \bottomrule
    \end{tabular}}
    \vspace{-5mm}
    \end{table}

\begin{wraptable}{r}{0.5\linewidth}
    \caption{Robustness evaluation against Auto-Attack and PGD $(k=100)$ with
    WRN-34-10 on the full test set of CIFAR10 dataset. ({*})
    Omit the cross-entropy loss of natural images. Detail can be found
    in Appendix \ref{sec:Exp-Setting}.\label{tab:res-wrn}}
    \centering
    \resizebox{0.9\linewidth}{!}{%
        \begin{tabular}{lcccc}
            \hline 
                & Nat & PGD & AA & C\&W\tabularnewline
            \hline 
            PGD-AT{*} & 84.93 & 55.04 & 52.12 & 40.85\tabularnewline
            UDR-PGD{*} & 84.60 & 55.71 & 52.98 & 47.31\tabularnewline
            \hline 
            TRADES & 85.70 & 56.97 & 53.82 & 47.65\tabularnewline
            UDR-TRADES & 84.93 & 57.35 & 54.45 & 49.14\tabularnewline
            \hline 
            AWP-AT & 85.57 & 57.78 & 53.91 & 49.91\tabularnewline
            UDR-AWP-AT & 85.51 & 58.65 & 54.40 & 54.44\tabularnewline
            \hline 
            \citet{zhang2020attacks} & 84.52 & - & 53.51 & - \tabularnewline
            \citet{huang2020self} & 83.48 & - & 53.34 & -\tabularnewline
            \citet{trades} & 84.92 & - & 53.08 & -\tabularnewline
            \citet{cui2020learnable} & 88.22 & - & 52.86 & -\tabularnewline
            \hline 
            \end{tabular}     
    }
\end{wraptable}

\textbf{Blackbox Attacks.~}
To further exam the generalization of the UDR framework, we conduct
the experiment with the blackbox setting via transferred attacks.
Specifically, we use PGD to generate adversarial examples according
to the model trained with a specific AT method, i.e., the \textit{source}
method. Next, we use the generated adversarial examples to attack
another AT method, i.e., the \textit{target} method. This is to see
whether an AT method can defend against attacks generated from other
models. We report the results in Table~\ref{tab:blackbox}. It can
be seen that with better generalization capacity, our UDR methods
also outperform their standard counterparts with a margin of 2\% 
in the blackbox setting.

\textbf{Results with WideResNet architecture.~}
We would like to provide further experimental results on the CIFAR10
dataset with WideResNet (WRN-34-10) as shown in Table \ref{tab:res-wrn}.
It can be seen that our distributional frameworks consistently outperform
their standard AT counterparts in both metrics. More specifically,
our improvement over PGD-AT against Auto-Attack is around 0.8\%, while
that for TRADES is 0.5\%. To make a more concrete conclusion, we deploy
our framework on a recent SOTA standard AT which is AWP-AT \citet{wu2020adversarial}.
The result shows that our distributional robustness version (UDR-AWP-AT)
also improves its counterpart by 0.5\%. With the same setting (i.e.,
same architecture and without additional data), our UDR-TRADES and
UDR-AWP-AT achieve better robustness than recently listed methods
on RobustBench \citep{croce2020robustbench}.\footnote{RobustBench reported a robust accuracy of 56.17\% for AWP-TRADES version
from \citet{wu2020adversarial} which is higher than ours but might
not be used as a reference.} 
Remarkably, the additional experiment with C\&W (L2) attack shows a significant improvement of our distributional methods over standard AT by around 5\%. 
More discussion can be found in Appendix \ref{sec:supp-exp}.

\subsection{Analytical Results}


\textbf{Benefit of the soft-ball projection.~}
Here we would like to analytically study why our UDR methods are better
than standard AT methods, by taking UDR-PGD and PGD-AT as examples. 
The visualization on the synthetic dataset can be found
in Appendix \ref{sec:exp-toy}.
Recall that one of the key differences between UDR-PGD and PGD-AT
is that the former uses the soft-ball projection and the later use
the hard-ball one, discussed in the second paragraph under Eq. (\ref{eq:find-xa}).
More specifically, Table \ref{tab:norm-perturbation} reports the average norm ($L_{1}$ and $L_{\infty}$)
of the perturbation $\delta=\left|x^{a}-x\right|_{p}$ in PGD and
our UDR-PGD. It can be seen that: (i) At the beginning of the training
process, there is no difference between the norms of the perturbations
generated by PGD and our UDR-PGD. More specifically, most of the pixels
lie on the edge of the hard-ball projection (i.e., $p(0.9\epsilon\leq\delta\leq\epsilon)=p(\delta\leq\epsilon)-p(\delta\leq0.9\epsilon)>80\%$).
(ii) When our model converges, there are 77.9\% pixels lying slightly
beyond the hard-ball projection (i.e., $p(\delta>\epsilon)$). It
is because our soft-ball projection can be adaptive based on the value
of . This flexibility helps the adversarial examples reach a better
local optimum of the prediction loss, therefore, benefits the adversarial
training. 

\begin{table}

\caption{Average norm $L_{1}$ and $L_{\infty}$ of the perturbation $\delta=\left|x^{a}-x\right|_{p}$}
\label{tab:norm-perturbation}
\centering
\resizebox{0.8\linewidth}{!}{%
    \begin{tabular}{lccccc}
    \hline 
        & $L_{1}$ & $L_{\infty}$ & $p(\delta\leq0.9\epsilon)$ & $p(\delta\leq\epsilon)$ & $p(\delta\leq1.1\epsilon)$\tabularnewline
    \hline 
    PGD & 0.0270 & 0.031 & 19.7\% & 100\% & 100\%\tabularnewline
    UDR-PGD at epoch 0th & 0.0278 & 0.031 & 18.9\% & 100\% & 100\%\tabularnewline
    UDR-PGD at epoch 200th & 0.0301 & 0.034 & 19.5\% & 22.1\% & 100\%\tabularnewline
    \hline 
    \end{tabular}
}

\end{table}

Next, we show that doing PGD adversarial training with larger $\epsilon$ cannot
achieve the same defence performance as our methods with the soft-ball
projection. We conduct more experiments with PGD-AT with $\epsilon=0.034$
(the final when our model converages) and $\epsilon=0.037$ to show
that simply extending the hard-ball projection doesn’t benefit adversarial
training. More specifically, the average robustness improvement with
$\epsilon=0.034$ is 0.75\%, while there is no improvement with $\epsilon=0.037$.

\begin{table}
\caption{Comparison to PGD-AT with different perturbation limitations.\label{tab:pgd-at-diff-eps}}

\centering{}%
\resizebox{0.7\linewidth}{!}{
    \begin{tabular}{lccccccc}
        \hline 
         & $\frac{8}{255}$ & $\frac{10}{255}$ & $\frac{12}{255}$ & $\frac{14}{255}$ & $\frac{16}{255}$ & $\frac{20}{255}$ & Avg\tabularnewline
        \hline 
        PGD-AT at $\epsilon=0.031$ & 46.0 & 33.7 & 23.7 & 15.2 & 9.5 & 3.6 & -\tabularnewline
        PGD-AT at $\epsilon=0.034$ & 46.7 & 34.8 & 24.7 & 16.2 & 10.1 & 3.7 & 0.75\tabularnewline
        PGD-AT at $\epsilon=0.037$ & 44.9 & 33.3 & 23.7 & 15.6 & 10.0 & 3.8 & -0.07\tabularnewline
        UDR-PGD at $\epsilon=0.031$ & 48.9 & 36.4 & 26.3 & 18.5 & 13.0 & 7.1 & 3.08\tabularnewline
        \hline 
        \end{tabular}
}
\vspace{-5mm}
\end{table}

\textbf{Parameter sensitivity of $\tau$.}
Figure~\ref{fig:fa}a and~\ref{fig:fa}b show the our framework's
sensitivity to $\tau$ on CIFAR10 under the PGD attack. It can be
observed that overly small values of $\tau$ can hardly improve adversarial
robustness while overly big values of $\tau$ may hurt the natural
performance ($acc_{nat}=68.7\%$ with $\tau=1.0$). Empirically, we
find that $\tau=2\eta$ performs well in our experiments.

\vspace{-4mm}
\section{Conclusions}

In this paper, we have presented a new unified distributional robustness
framework for adversarial training, which unifies and generalizes
standard AT approaches with improved adversarial robustness. By defining
a new family of risk functions, our framework facilitates the development
of the distributional robustness counterparts of the SOTA AT methods
including PGD-AT, TRADES, MART and AWP. Moreover, we introduce a new cost
function, which enables us to bridge the connections between standard
AT methods and their distributional robustness counterparts and to
show that the former ones can be viewed as the special cases of the
later ones. Extensive experiments on the benchmark datasets including
MNIST, CIFAR10, CIFAR100 show that our proposed algorithms are able to boost
the model robustness against strong attacks with better generalization
capacity.

\section*{Acknowledgement}
This work was partially supported by the Australian Defence Science and Technology (DST) Group under the Next Generation Technology Fund (NGTF) scheme. 
The authors are grateful to the anonymous (meta) reviewers for
their helpful comments. 
\bibliography{iclr2022_conference}
\bibliographystyle{iclr2022_conference}

\appendix

\section{Theoretical Development \label{sec:Proof}}


\textbf{Theorem 1.} \emph{With the cost function $\tilde{c}_{\mathcal{X}}$
defined as above , the optimization problem: 
\begin{equation}
\inf_{\theta,\lambda\geq0}\left\{ \lambda\epsilon+\mathbb{E}_{\mathbb{P}}\left[\sup_{x'}\left\{ g_{\theta}\left(x',x,y\right)-\lambda\tilde{c}_{\mathcal{X}}\left(x',x\right)\right\} \right]\right\} \label{eq:udr_unified}
\end{equation}
is equivalent to the optimization problem: 
\begin{equation}
\inf_{\theta}\mathbb{E}_{\mathbb{P}}\left[\sup_{x'\in B_{\epsilon}\left(x\right)}g_{\theta}\left(x',x,y\right)\right].
\end{equation}
}
\begin{proof}
We need to prove that 
\begin{equation}
\inf_{\lambda\geq0}\left\{ \lambda\epsilon+\mathbb{E}_{\mathbb{P}}\left[\sup_{x'}\left\{ g_{\theta}\left(x',x,y\right)-\lambda\tilde{c}_{\mathcal{X}}\left(x',x\right)\right\} \right]\right\} =\mathbb{E}_{\mathbb{P}}\left[\sup_{x'\in B_{\epsilon}\left(x\right)}g_{\theta}\left(x',x,y\right)\right].\label{eq:left_right}
\end{equation}

By the definition of the cost function $\tilde{c}_{\mathcal{X}}$,
the LHS of (\ref{eq:left_right}) can be rewritten as:{\small{}{}
\begin{equation}
\min\left\{ \inf_{\lambda>0}\left\{ \lambda\epsilon+\mathbb{E}_{\mathbb{P}}\left[\sup_{x'\in B_{\epsilon}\left(x\right)}\left\{ g_{\theta}\left(x',x,y\right)-\lambda c_{\mathcal{X}}\left(x',x\right)\right\} \right]\right\} ,\mathbb{E}_{\mathbb{P}}\left[\sup_{x'}g_{\theta}\left(x',x,y\right)\right]\right\} .\label{eq:unified}
\end{equation}
}{\small\par}

Given any $\lambda>0$ and $x'\in B_{\epsilon}\left(x\right)$, we
have 
\begin{gather*}
\lambda\epsilon+g_{\theta}\left(x',x,y\right)-\lambda c_{\mathcal{X}}\left(x',x\right)=g_{\theta}\left(x',x,y\right)+\lambda\left(\epsilon-c_{\mathcal{X}}\left(x',x\right)\right)\geq\mathbb{E}_{\mathbb{P}}\left[g_{\theta}\left(x',x,y\right)\right].
\end{gather*}

Hence, we arrive at 
\begin{gather*}
\lambda\epsilon+\sup_{x'\in B_{\epsilon}\left(x\right)}\left\{ g_{\theta}\left(x',x,y\right)-\lambda c_{\mathcal{X}}\left(x',x\right)\right\} \geq\sup_{x'\in B_{e}\left(x\right)}g_{\theta}\left(x',x,y\right).
\end{gather*}
\[
\lambda\epsilon+\mathbb{E}_{\mathbb{P}}\left[\sup_{x'\in B_{\epsilon}\left(x\right)}\left\{ g_{\theta}\left(x',x,y\right)-\lambda c_{\mathcal{X}}\left(x',x\right)\right\} \right]\geq\mathbb{E}_{\mathbb{P}}\left[\sup_{x'\in B_{e}\left(x\right)}g_{\theta}\left(x',x,y\right)\right].
\]
which follows that 
\begin{gather}
\inf_{\lambda>0}\left\{ \lambda\epsilon+\mathbb{E}_{\mathbb{P}}\left[\sup_{x'\in B_{\epsilon}\left(x\right)}\left\{ g_{\theta}\left(x',x,y\right)-\lambda c_{\mathcal{X}}\left(x',x\right)\right\} \right]\right\} \nonumber \\
\geq\mathbb{E}_{\mathbb{P}}\left[\sup_{x'\in B_{e}\left(x\right)}\mathbb{E}_{\mathbb{P}}\left[g_{\theta}\left(x',x,y\right)\right]\right].\label{eq:inequality}
\end{gather}

We now prove the inequality 
\begin{gather*}
\lim_{\lambda\rightarrow0^{+}}\left\{ \lambda\epsilon+\mathbb{E}_{\mathbb{P}}\left[\sup_{x'\in B_{\epsilon}\left(x\right)}\left\{ g_{\theta}\left(x',x,y\right)-\lambda c_{\mathcal{X}}\left(x',x\right)\right\} \right]\right\} \\
=\mathbb{E}_{\mathbb{P}}\left[\sup_{x'\in B_{e}\left(x\right)}\mathbb{E}_{\mathbb{P}}\left[g_{\theta}\left(x',x,y\right)\right]\right].
\end{gather*}

Take a sequence $\left\{ \lambda_{n}\right\} _{n\geq1}\rightarrow0^{+}$.
Given a feasible pair $\left(x,y\right)$, we define 
\[
f_{n}\left(x';x,y\right):=g_{\theta}\left(x',x,y\right)+\lambda_{n}\left[\epsilon-c_{\mathcal{X}}\left(x',x\right)\right],\forall x'\in B_{\epsilon}\left(x\right).
\]

It is evident that $f_{n}\left(x';x,y\right)$ converges pointwise
to $g_{\theta}\left(x',x,y\right)$ over the compact set $B_{\epsilon}\left(x\right)$.
Therefore, $f_{n}\left(x';x,y\right)$ converges uniformly to $g_{\theta}\left(x',x,y\right)$
on this set. This follows that 
\[
\forall\alpha>0,\exists n_{0}=n\left(\alpha\right):\left|f_{n}\left(x';x,y\right)-g_{\theta}\left(x',x,y\right)\right|<\alpha,\forall x'\in B_{\epsilon}\left(x\right),n\geq n_{0}.
\]

Hence, we obtain for all $x'\in B_{\epsilon}\left(x\right)$ and $n\geq n_{0}$:
\[
g_{\theta}\left(x',x,y\right)-\alpha<f_{n}\left(x';x,y\right)<g_{\theta}\left(x',x,y\right)+\alpha.
\]

This leads to the following for all $n\geq n_{0}$: 
\[
\sup_{x'\in B_{\epsilon}\left(x\right)}g_{\theta}\left(x',x,y\right)-\alpha\leq\sup_{x'\in B_{\epsilon}\left(x\right)}f_{n}\left(x';x,y\right)\leq\sup_{x'\in B_{\epsilon}\left(x\right)}g_{\theta}\left(x',x,y\right)+\alpha.
\]

Therefore, we obtain: 
\[
\lim_{n\rightarrow\infty}\sup_{x'\in B_{\epsilon}\left(x\right)}f_{n}\left(x';x,y\right)=\sup_{x'\in B_{\epsilon}\left(x\right)}g_{\theta}\left(x',x,y\right)
\]
for all feasible pairs $\left(x,y\right)$, which further means that
\[
\lim_{n\rightarrow\infty}\mathbb{E}_{\mathbb{P}}\left[\sup_{x'\in B_{\epsilon}\left(x\right)}f_{n}\left(x';x,y\right)\right]=\mathbb{E}_{\mathbb{P}}\left[\sup_{x'\in B_{\epsilon}\left(x\right)}g_{\theta}\left(x',x,y\right)\right],
\]
or equivalently 
\begin{equation}
\lim_{n\rightarrow\infty}\mathbb{E}_{\mathbb{P}}\left[\lambda_{n}\epsilon+\mathbb{E}_{\mathbb{P}}\left[\sup_{x'\in B_{\epsilon}\left(x\right)}\left\{ g_{\theta}\left(x',x,y\right)-\lambda_{n}c_{\mathcal{X}}\left(x',x\right)\right\} \right]\right]=\mathbb{E}_{\mathbb{P}}\left[\sup_{x'\in B_{\epsilon}\left(x\right)}g_{\theta}\left(x',x,y\right)\right].\label{eq:lim_seq}
\end{equation}

Because Eq. (\ref{eq:lim_seq}) holds for every sequence $\left\{ \lambda_{n}\right\} _{n\geq1}\rightarrow0^{+}$,
we reach 
\begin{align}
 & \lim_{\lambda\rightarrow0^{+}}\left\{ \lambda\epsilon+\mathbb{E}_{\mathbb{P}}\left[\sup_{x'\in B_{\epsilon}\left(x\right)}\left\{ g_{\theta}\left(x',x,y\right)-\lambda c_{\mathcal{X}}\left(x',x\right)\right\} \right]\right\} \nonumber \\
 & =\mathbb{E}_{\mathbb{P}}\left[\sup_{x'\in B_{e}\left(x\right)}\mathbb{E}_{\mathbb{P}}\left[g_{\theta}\left(x',x,y\right)\right]\right].\label{eq:lim_all}
\end{align}

By combining (\ref{eq:inequality}) and (\ref{eq:lim_all}), we reach
\begin{gather}
\inf_{\lambda>0}\left\{ \lambda\epsilon+\mathbb{E}_{\mathbb{P}}\left[\sup_{x'\in B_{\epsilon}\left(x\right)}\left\{ g_{\theta}\left(x',x,y\right)-\lambda c_{\mathcal{X}}\left(x',x\right)\right\} \right]\right\} \nonumber \\
=\mathbb{E}_{\mathbb{P}}\left[\sup_{x'\in B_{e}\left(x\right)}\mathbb{E}_{\mathbb{P}}\left[g_{\theta}\left(x',x,y\right)\right]\right].\label{eq:equality}
\end{gather}

Finally, we have 
\begin{align*}
 & \inf_{\lambda\geq0}\left\{ \lambda\epsilon+\mathbb{E}_{\mathbb{P}}\left[\sup_{x'}\left\{ g_{\theta}\left(x',x,y\right)-\lambda\tilde{c}_{\mathcal{X}}\left(x',x\right)\right\} \right]\right\} \\
=\min & \left\{ \inf_{\lambda>0}\left\{ \lambda\epsilon+\mathbb{E}_{\mathbb{P}}\left[\sup_{x'\in B_{\epsilon}\left(x\right)}\left\{ g_{\theta}\left(x',x,y\right)-\lambda c_{\mathcal{X}}\left(x',x\right)\right\} \right]\right\} ,\mathbb{E}_{\mathbb{P}}\left[\sup_{x'}g_{\theta}\left(x',x,y\right)\right]\right\} \\
= & \min\left\{ \mathbb{E}_{\mathbb{P}}\left[\sup_{x'\in B_{e}\left(x\right)}\mathbb{E}_{\mathbb{P}}\left[g_{\theta}\left(x',x,y\right)\right]\right],\mathbb{E}_{\mathbb{P}}\left[\sup_{x'}g_{\theta}\left(x',x,y\right)\right]\right\} \\
 & =\mathbb{E}_{\mathbb{P}}\left[\sup_{x'\in B_{e}\left(x\right)}\mathbb{E}_{\mathbb{P}}\left[g_{\theta}\left(x',x,y\right)\right]\right].
\end{align*}

That concludes our proof. 
\end{proof}
One of most technical challenge we need to bypass in our work is that
in theory developed in \citet{blanchet2019quantifying}, to equivalently
transform the primal form to the dual form, it requires the cost function
to be finite. In the following theorem, we reprove the equivalence
of the primal and dual forms in our context. 
\begin{thm}
\label{thm:primal_dual_form}Assume that the function $g$ is upper-bounded
by a number $L$. We have the following equality between the primal
form and dual form 
\[
\sup_{\mathbb{Q}:\mathcal{W}_{c}\left(\mathbb{Q},\mathbb{P}_{\triangle}\right)<\epsilon}\mathbb{E}_{\mathbb{Q}}\left[g\left(z'\right)\right]=\inf_{\lambda\geq0}\left\{ \lambda\epsilon+\mathbb{E}_{\mathbb{P}_{\triangle}}\left[\sup_{z'}\left\{ g\left(z'\right)-\lambda c\left(z',z\right)\right\} \right]\right\} ,
\]
where $z=\left(x,x,y\right)$, $z'=\left(x',x'',y'\right)$, and we
have defined 
\[
c\left(z,z'\right)=\tilde{c}_{\mathcal{X}}\left(x,x'\right)+\infty\times\tilde{c}_{\mathcal{X}}\left(x,x''\right)+\infty\times\boldsymbol{1}\left\{ y\neq y'\right\} ,
\]
for which we have defined 
\[
\tilde{c}_{\mathcal{X}}\left(x,x'\right)=\begin{cases}
c_{\mathcal{X}}\left(x,x'\right) & \text{if}\,c_{\mathcal{X}}(x,x')\leq\epsilon\\
\infty & \text{otherwise}.
\end{cases}
\]
\end{thm}

\begin{proof}
Given a positive integer number $n>0$, we define the following metrics:
\[
c^{n}\left(z,z'\right)=\tilde{c}_{\mathcal{X}}^{n}\left(x,x'\right)+\infty\times\tilde{c}_{\mathcal{X}}^{n}\left(x,x''\right)+\infty\times\boldsymbol{1}\left\{ y\neq y'\right\} ,
\]
\[
\tilde{c}_{\mathcal{X}}^{n}\left(x,x'\right)=\begin{cases}
c_{\mathcal{X}}\left(x,x'\right) & \text{if}\,c_{\mathcal{X}}\left(x,x'\right)<\epsilon.\\
n & \text{otherwise}.
\end{cases}
\]

We have $\tilde{c}_{\mathcal{X}}^{n}\nearrow\tilde{c}_{\mathcal{X}}$
and $c^{n}\nearrow c$. We now prove that 
\[
\sup_{\mathbb{Q:\mathcal{W}_{c}}\left(\mathbb{Q},\mathbb{P}_{\triangle}\right)<\epsilon}\mathbb{E}_{\mathbb{Q}}\left[g\left(z'\right)\right]=\inf_{n}\sup_{\mathbb{Q}:\mathcal{W}_{c^{n}}\left(\mathbb{Q},\mathbb{P}_{\triangle}\right)<\epsilon}\mathbb{E}_{\mathbb{Q}}\left[g\left(z'\right)\right].
\]

In fact, for each $n$, we have $c^{n}\leq c$. Therefore, $\mathcal{W}_{c^{n}}\left(\mathbb{Q},\mathbb{P}_{\triangle}\right)\leq\mathbb{\mathcal{W}}_{c}\left(\mathbb{Q},\mathbb{P}_{\triangle}\right)$,
hence $\left\{ \mathbb{Q:\mathcal{W}}_{c}\left(\mathbb{Q},\mathbb{P}_{\triangle}\right)<\epsilon\right\} \subset\left\{ \mathbb{Q}:\mathcal{W}_{c^{n}}\left(\mathbb{Q},\mathbb{P}_{\triangle}\right)<\epsilon\right\} $,
implying that 
\[
\sup_{\mathbb{Q:\mathcal{W}_{c}}\left(\mathbb{Q},\mathbb{P}_{\triangle}\right)<\epsilon}\mathbb{E}_{\mathbb{Q}}\left[g\left(z'\right)\right]\leq\sup_{\mathbb{Q}:\mathcal{W}_{c^{n}}\left(\mathbb{Q},\mathbb{P}_{\triangle}\right)<\epsilon}\mathbb{E}_{\mathbb{Q}}\left[g\left(z'\right)\right].
\]

\[
\sup_{\mathbb{Q:\mathcal{W}_{c}}\left(\mathbb{Q},\mathbb{P}_{\triangle}\right)<\epsilon}\mathbb{E}_{\mathbb{Q}}\left[g\left(z'\right)\right]\leq\inf_{n}\sup_{\mathbb{Q}:\mathcal{W}_{c^{n}}\left(\mathbb{Q},\mathbb{P}_{\triangle}\right)<\epsilon}\mathbb{E}_{\mathbb{Q}}\left[g\left(z'\right)\right].
\]

Let us define 
\[
A=\cup_{\left(x,y\right)\in\mathcal{D}}\left\{ \left(z,z'\right):z=\left(x,x,y\right),z'=\left(x',x'',y'\right),c_{\mathcal{X}}\left(x,x'\right)<\epsilon,x''=x,y'=y\right\} ,
\]

\[
B=\cup_{\left(x,y\right)\in\mathcal{D}}\left\{ \left(z,z'\right):z=\left(x,x,y\right),z'=\left(x',x'',y'\right),c_{\mathcal{X}}\left(x,x'\right)\geq\epsilon,x''=x,y'=y\right\} .
\]

To simplify our proof, without generalization ability, for each $n$,
we denote $\mathbb{Q}_{n}$ as the distribution in $\left\{ \mathbb{Q}:\mathcal{W}_{c^{n}}\left(\mathbb{Q},\mathbb{P}_{\triangle}\right)<\epsilon\right\} $
that peaks $\mathbb{E}_{\mathbb{Q}}\left[g_{\theta}\left(z'\right)\right]$
and $\gamma_{n}$ as the optimal transport plan of $\mathcal{W}_{c^{n}}\left(\mathbb{Q}_{n},\mathbb{P}_{\triangle}\right)$
which admits $\mathbb{P}_{\triangle}$ and $\mathbb{Q}_{n}$ as its
marginals. Note that because $\mathcal{W}_{c^{n}}\left(\mathbb{Q}_{n},\mathbb{P}_{\triangle}\right)<\epsilon$,
the support of $\gamma_{n}$ almost surely determines on $A\cup B$.We
then have 
\begin{align*}
\mathcal{W}_{c^{n}}\left(\mathbb{Q}_{n},\mathbb{P}_{\triangle}\right) & =\int c^{n}\left(z,z'\right)d\gamma_{n}\left(z,z'\right)\\
= & \int_{A}c^{n}\left(z,z'\right)d\gamma_{n}\left(z,z'\right)+\int_{B}c^{n}\left(z,z'\right)d\gamma_{n}\left(z,z'\right)\\
= & \int_{A}c_{\mathcal{X}}\left(x,x'\right)d\gamma_{n}\left(z,z'\right)+\int_{B}nd\gamma_{n}\left(z,z'\right)\\
= & \int_{A}c_{\mathcal{X}}\left(x,x'\right)d\gamma_{n}\left(z,z'\right)+n\gamma_{n}\left(B\right)<\epsilon.
\end{align*}

Therefore, we obtain: $\gamma_{n}\left(B\right)<\frac{\epsilon}{n}$.
We now define $\bar{\gamma}_{n}$ as a restricted measure of $\gamma_{n}$
on $A$, meaning that $\bar{\gamma}_{n}\left(C\right)=\frac{\gamma_{n}\left(A\right)+\gamma_{n}\left(B\right)}{\gamma_{n}\left(A\right)}\gamma_{n}\left(C\right)=\left(1+o\left(n^{-1}\right)\right)\gamma_{n}\left(C\right)$
for any measure set $C\subset A$, where $\lim_{n\rightarrow\infty}o\left(n^{^{-1}}\right)=0$.
Let $\mathbb{P}_{n}$ as marginal distribution of $\mathbb{Q}_{n}$
corresponding to the dimensions of $z'$. It appears that 
\begin{align*}
\mathcal{W}_{c}\left(\mathbb{P}_{n},\mathbb{P}_{\triangle}\right) & \leq\int_{A}c\left(z,z'\right)d\bar{\gamma}_{n}\left(z,z'\right)+\int_{B}c\left(z,z'\right)d\bar{\gamma}_{n}\left(z,z'\right)\\
 & \overset{(1)}{=}\int_{A}c_{\mathcal{X}}\left(x,x'\right)d\bar{\gamma}_{n}\left(z,z'\right)<\int_{A}\epsilon d\bar{\gamma}_{n}\left(z,z'\right)=\epsilon.
\end{align*}

Note that we have $\overset{(1)}{=}$ because $\bar{\gamma}_{n}\left(B\right)=0$.

This implies that $\mathbb{P}_{n}\in\left\{ \mathbb{Q:\mathcal{W}}_{c}\left(\mathbb{Q},\mathbb{P}_{\triangle}\right)<\epsilon\right\} $,
which follows that 
\begin{align*}
\sup_{\mathbb{Q:\mathcal{W}_{c}}\left(\mathbb{Q},\mathbb{P}_{\triangle}\right)<\epsilon}\mathbb{E}_{\mathbb{Q}}\left[g\left(z'\right)\right] & \geq\mathbb{E}_{\mathbb{\mathbb{P}}_{n}}\left[g_{\theta}\left(z'\right)\right]=\mathbb{E}_{\bar{\gamma}_{n}}\left[g\left(z'\right)\right]\\
= & \int_{A}g\left(z'\right)d\bar{\gamma}_{n}\left(z,z'\right)+\int_{B}g\left(z'\right)d\bar{\gamma}_{n}\left(z,z'\right)\\
\overset{(1)}{=} & \int_{A}g\left(z'\right)d\bar{\gamma}_{n}\left(z,z'\right)=\frac{\gamma_{n}\left(A\right)+\gamma_{n}\left(B\right)}{\gamma_{n}\left(A\right)}\int_{A}g\left(z'\right)d\gamma_{n}\left(z,z'\right)\\
= & \left(1+o\left(n^{^{-1}}\right)\right)\left[\int_{A\cup B}g\left(z'\right)d\gamma_{n}\left(z,z'\right)-\int_{B}g\left(z'\right)d\gamma_{n}\left(z,z'\right)\right]\\
= & \left(1+o\left(n^{^{-1}}\right)\right)\left[\int_{A\cup B}g\left(z'\right)d\mathbb{Q}_{n}\left(z'\right)-\int_{B}g\left(z'\right)d\gamma_{n}\left(z,z'\right)\right]\\
\geq & \left(1+o\left(n^{^{-1}}\right)\right)\left[\sup_{\mathbb{Q}:\mathcal{W}_{c^{n}}\left(\mathbb{Q},\mathbb{P}_{\triangle}\right)<\epsilon}\mathbb{E}_{\mathbb{Q}}\left[g_{\theta}\left(z'\right)\right]-\int_{B}Ld\gamma_{n}\left(z,z'\right)\right]
\end{align*}
\begin{align*}
\sup_{\mathbb{Q:\mathcal{W}_{c}}\left(\mathbb{Q},\mathbb{P}_{\triangle}\right)<\epsilon}\mathbb{E}_{\mathbb{Q}}\left[g\left(z'\right)\right] & \geq\left(1+o\left(n^{^{-1}}\right)\right)\left[\sup_{\mathbb{Q}:\mathcal{W}_{c^{n}}\left(\mathbb{Q},\mathbb{P}_{\triangle}\right)<\epsilon}\mathbb{E}_{\mathbb{Q}}\left[g_{\theta}\left(z'\right)\right]-L\gamma_{n}\left(B\right)\right]\\
 & \overset{(2)}{\geq}\left(1+o\left(n^{^{-1}}\right)\right)\left[\sup_{\mathbb{Q}:\mathcal{W}_{c^{n}}\left(\mathbb{Q},\mathbb{P}_{\triangle}\right)<\epsilon}\mathbb{E}_{\mathbb{Q}}\left[g_{\theta}\left(z'\right)\right]-\frac{L\epsilon}{n}\right].
\end{align*}

Note that we have $\overset{(1)}{=}$ due to $\bar{\gamma}_{n}\left(B\right)=0$
and $\overset{(2)}{\geq}$ due to $\gamma_{n}\left(B\right)<\frac{\epsilon}{n}$.
Therefore, we reach the conclusion 
\[
\sup_{\mathbb{Q:\mathcal{W}_{c}}\left(\mathbb{Q},\mathbb{P}_{\triangle}\right)<\epsilon}\mathbb{E}_{\mathbb{Q}}\left[g_{\theta}\left(z'\right)\right]=\inf_{n}\sup_{\mathbb{Q}:\mathcal{W}_{c^{n}}\left(\mathbb{Q},\mathbb{P}_{\triangle}\right)<\epsilon}\mathbb{E}_{\mathbb{Q}}\left[g_{\theta}\left(z'\right)\right].
\]

Next, we apply primal-dual form in \citet{blanchet2019quantifying}
for the finite metric $\tilde{c}_{\mathcal{X}}^{n}$ to reach 
\[
\sup_{\mathcal{W}_{c^{n}}\left(\mathbb{Q},\mathbb{P}_{\triangle}\right)<\epsilon}\mathbb{E}_{\mathbb{Q}}\left[g_{\theta}\left(z'\right)\right]=\inf_{\lambda\geq0}\left\{ \lambda\epsilon+\mathbb{E}_{\mathbb{P}_{\triangle}}\left[\sup_{z'}\left\{ g_{\theta}\left(z'\right)-\lambda c^{n}\left(z',z\right)\right\} \right]\right\} .
\]
Finally, taking $n\rightarrow\infty$ and noting that $c^{n}\nearrow c$,
we reach the conclusion. 
\end{proof}

\section{Further Explanation Why Our UDR can utilize Global Information and
the Advantage of Soft-ball\label{sec:global_information}}

\begin{algorithm}
\textbf{Input}: training set $\mathcal{D}$, number of iterations
$T$, batch size $N$, adversary parameters $\{k,\epsilon,\eta\}$

\textbf{for }$t=1$ to $T$ \textbf{do} 
\begin{enumerate}
\item \textbf{Sample} mini-batch $\{x_{i},y_{i}\}_{i=1}^{N}\sim\mathcal{D}$ 
\item \textbf{Find} adversarial examples $\{x_{i}^{a}\}_{i=1}^{N}$ using
Eq. (\ref{eq:find-xa}) 
\begin{enumerate}
\item \textbf{Initialize} randomly: $x_{i}^{0}=x_{i}+noise$ where $noise\sim\mathcal{U}(-\epsilon,\epsilon)$ 
\item \textbf{for} $n=1$ to $k$ \textbf{do} 
\begin{enumerate}
\item $x_{i}^{inter}=x_{i}^{n}+\eta\text{sign}\left(\nabla_{x}g_{\theta}(x_{i}^{n},x_{i},y_{i})\right)$ 
\item $x_{i}^{n+1}=x_{i}^{inter}-\eta\lambda\nabla_{x}\hat{c}(x_{i}^{inter},x_{i})$ 
\end{enumerate}
\item \textbf{Clip} to valid range: $x_{i}^{a}=clip(x_{i}^{k},0,1)$ 
\end{enumerate}
\item \textbf{Update} parameter $\lambda$ using Eq. (\ref{eq:update-lamda}) 
\item \textbf{Update} model parameter $\theta$ using Eq. (\ref{eq:update-theta}) 
\end{enumerate}
\textbf{Output: }model parameter\textbf{ $\theta$}

\caption*{\textbf{Algorithm 1} The pseudocode of our proposed method.}
\end{algorithm}

The advantage of our soft ball comes from the adaptive capability
of $\lambda$, which is controlled by a global effect regarding how
far adversarial examples $x_{i}^{a}$ from benign examples $x_{i}$.
Let us revisit Algorithm 1. In the step 2.(b).i, we update 
\[
x_{i}^{inter}=x_{i}^{n}+\eta\text{sign}\left(\nabla_{x}g_{\theta}(x_{i}^{n},x_{i},y_{i})\right)
\]
 with the aim to find $x_{i}^{inter}$ that can maximize $g_{\theta}(\cdot,x_{i},y_{i})$
as in the standard versions.

Furthermore, in the step 2.(b).ii, we update 
\begin{align}
x_{i}^{n+1} & =x_{i}^{inter}-\eta\lambda\nabla_{x}\hat{c}(x_{i}^{inter},x_{i})=x_{i}^{inter}-\eta\lambda\left(x_{i}^{inter}-x_{i}\right) \nonumber \\
 & =\left(1-\eta\lambda\right)x_{i}^{inter}+\eta\lambda x_{i},
\label{eq:update-x-adv}
\end{align}
where we assume L2 cost $c(x,x')=\frac{1}{2}\Vert x-x'\Vert^{2}$
is used. It is evident that $x_{i}^{n+1}$ is an interpolation point
of $x_{i}^{inter}$ and $x_{i}$, hence $x_{i}^{n+1}$ is drawn back
to $x_{i}$ wherein the drawn-back amount is proportional to $\eta\lambda$. 

We now revisit the formula to update $\lambda$ as Eq. (\ref{eq:update-lamda})
\begin{align*}
\lambda_{n} & =\lambda-\eta_{\lambda}\left(\epsilon-\frac{1}{N}\sum_{i=1}^{N}\hat{c}_{\mathcal{X}}(x_{i}^{a},x_{i})\right),
\end{align*}
which indicates that $\lambda$ is globally controlled. More specifically,
if average distance from $x_{i}^{a}$ to $x_{i}$ (i.e., $\frac{1}{N}\sum_{i=1}^{N}\hat{c}_{\mathcal{X}}(x_{i}^{a},x_{i})$)
is less than $\epsilon$ (i.e., adversarial examples are globally
close to benign examples), $\lambda$ is adapted decreasingly. Linking
with the formula in Eq. (\ref{eq:update-x-adv}), in this case, $x_{i}^{n+1}$ gets back
to $x_{i}$ less aggressively to maintain the distance between $x_{i}^{a}$
and $x_{i}$. Otherwise, adversarial examples are globally far from
benign examples, $\lambda$ is adapted increasingly. In this case,
$x_{i}^{n+1}$ gets back to $x_{i}$ more aggressively to reduce more
the distance between $x_{i}^{a}$ and $x_{i}$. 

\section{Related Work}

\textbf{Adversarial Attacks.} In this paper, we are interested in
image classification tasks and focus on the adversaries that add small
perturbations to the pixels of an image to generate attacks based
on gradients, which are the most popular and effective. FGSM \citep{goodfellow2014explaining}
and PGD \citep{madry2017towards} are the most representative gradient-based
attacks and  PGD is the most widely-used one, due to its effectiveness
and simplicity. Recently, there are several variants of PGD that achieve
improved performance, for example, Auto-Attack by ensembling PGD with
other attacks \citep{croce2020minimally} and the B\&B method \citep{brendel2019accurate}
by attacking with decision-based boundary initialized with PGD. Along
with PGD, these attacks have been considered as benchmark attacks
for adversarial robustness.

\textbf{Adversarial defenses.} Among various kinds of defense approaches,
Adversarial Training (AT), originating in \cite{goodfellow2014explaining},
has drawn the most research attention. Given its effectiveness and
efficiency, many variants of AT have been proposed with (1) different
types of adversarial examples (e.g., the worst-case examples as in \cite{goodfellow2014explaining}
or most divergent examples as in \cite{trades}), (2) different
searching strategies (e.g., non-iterative FGSM and Rand FGSM
\citep{madry2017towards}), (3) additional regularizations (e.g.,
adding constraints in the latent space \citep{zhang2019defense,bui2020improving,bui2021understanding,pmlr-v119-hoang20c}),
and (4) different model architectures (e.g., activation function \citep{xie2020smooth}
or ensemble models \citep{pang2019improving, bui2021improving}).

\textbf{Distributional robustness.}
There have been a few works attempting to connect DR with adversarial
machine learning or improve adversarial robustness based on the ideas
of DR \citep{sinha2017certifying,staib2017distributionally,miyato2018virtual,zhang2019defense,najafi2019robustness,levine2020wasserstein,le2022_global_local,nguyen-duc2022particle}.
A recent work
of \cite{deng2020adversarial} proposes a new AT algorithm by constructing
a distribution over each data sample to model the adversarial examples
around it, which is still in the category of pointwise adversary \citep{sinha2017certifying}
and has no relations to DR. Although its aim of enhancing adversarial
robustness is visually related ours, its mythology is different from
ours.
Therefore, we consider \cite{sinha2017certifying,staib2017distributionally}
as the most relevant ones to ours. Specifically, both works leverage
the dual form of Wasserstein DR \citep{blanchet2019quantifying} for
searching worst-case perturbations for AT, where \cite{sinha2017certifying}
(WRM) focuses on certified robustness with comprehensive study on
the tradeoffs between complexity, generality, guarantees, and speed,
while \cite{staib2017distributionally} (FDRO) points out that Wasserstein
robust optimization can be viewed as the generalization to standard
AT. 

Although our study is inspired by the two works, there are significant
differences and new results of ours: \textbf{1)} We introduce a new Wasserstein cost function and a new series of risk functions in WDR, which facilitate our framework to generalize and encompass many SOTA AT methods. While WRM can be viewed as the generalization to PGD-AT only.
\textbf{2)} Most importantly, although WDR has been demonstrated to have superior properties over standard AT in the two papers, unfortunately, WRM and FDRO have not been observed to outperform standard AT methods. For example, the experiments of FDRO show that adversarial robustness on MNIST of WRM and FDRO is worse than that of AT with PGD and iterative-FGSM~\citep{staib2017distributionally}. Moreover, WRM and FDRO's effectiveness either on more complex colored images (e.g., CIFAR10) or against more
advanced attacks (e.g., Auto-Attack) has not been carefully studied
yet. On the contrary, we conduct extensive experiments to show the
SOTA performance of our proposed algorithms. 

\section{Experimental Settings \label{sec:Exp-Setting}}

\paragraph{For MNIST dataset.}

We use a standard CNN architecture for the MNIST dataset which is
identical with that in \citet{carlini2017towards}. We use the SGD
optimizer with momentum 0.9, starting learning rate 1e-2 and reduce
the learning rate ($\times0.1$) at epoch \{55, 75, 90\}. We train
with 100 epochs.

\paragraph{For CIFAR10 and CIFAR100 dataset with ResNet18 architecture.}

We use the ResNet18 for the CIFAR10 and CIFAR100 dataset. We use the
SGD optimizer with momentum 0.9, weight decay 3.5e-3 as in the official implementation 
from \citet{wang2019improving}.\footnote{https://github.com/YisenWang/MART}
The starting learning rate 1e-2 and reduce the learning rate ($\times0.1$)
at epoch \{75, 90, 100\}. We train with 200 epochs.

\paragraph{For hard/soft-ball projection experiments. }

For PGD-AT, we use the following three ad-hoc strategies for $\epsilon$:
1) Fixing $\epsilon=8/255$; 2) Fixing $\epsilon=16/255$; 3) Gradually
increasing/decreasing $\epsilon$from $8/255$ to $16/255$, from
epoch 20 to epoch 70, with the changing rate $\delta=8/255/50$ per
epoch. For example, the perturbation bound of the increasing strategy
at epoch $i$ is: $\epsilon_{i}=\min(\frac{16}{255},\max(\frac{8}{255},\frac{8}{255}+(i-20)\delta))$;
the perturbation bound for decreasing strategy is: $\epsilon_{i}=\max(\frac{8}{255},\min(\frac{16}{255},\frac{16}{255}-(i-20)\delta))$.

\paragraph{For CIFAR10 with WideResNet architecture.}

We follow the setting in \citet{pang2020bag} for the additional
experiments on CIFAR10 with WideResNet-34-10 architecture. More specifically,
we train with 200 epochs with SGD optimizer with momentum 0.9, weight
decay 5e-4. The learning rate is 0.1 and reduce at epoch 100th and
150th with rate 0.1 \citep{rice2020overfitting,wu2020adversarial}. More importantly, to match the performance as
reported in \citet{croce2020robustbench}, we omit the cross-entropy
loss of the natural images in PGD-AT and UDR-PGD. More specifically,
the objective function of PGD-AT in Eq. (\ref{eq:pdg-at}) has been replaced by: $\inf_{\theta}\mathbb{E}_{\mathbb{P}}\left[\beta\sup_{x'\in B_{\epsilon}\left(x\right)}CE\left(h_{\theta}\left(x'\right),y\right)\right]$
while the unified risk function for UDR-PGD to be: $g_{\theta}\left(z'\right):=\beta CE\left(h_{\theta}\left(x'\right),y'\right)$. 
We also switch Batch Normalization layer to evaluation stage when crafting adversarial examples as adviced in \citet{pang2020bag}. 

\section{Visualizing the Benefit of Distributional Robustness\label{sec:exp-toy}}

\paragraph{Synthetic dataset setting.}

\begin{figure}[ht!]
\centering \begin{subfigure}[b]{0.25\textwidth} \centering \includegraphics[width=1\textwidth]{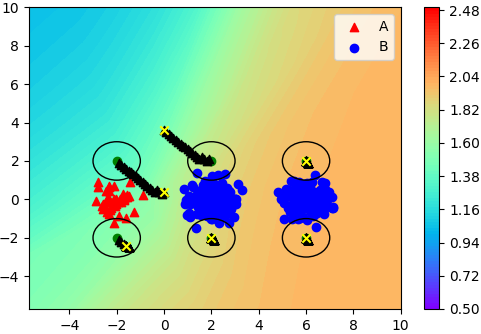}
\caption{UDR-PGD}
\label{fig:udr-pgd} \end{subfigure} \begin{subfigure}[b]{0.25\textwidth}
\centering \includegraphics[width=1\textwidth]{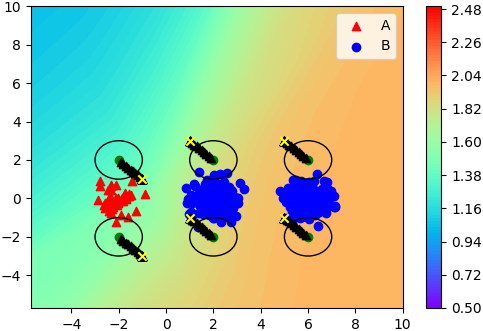}
\caption{PGD}
\label{fig:pgd-traject} \end{subfigure} \begin{subfigure}[b]{0.25\textwidth}
\centering \includegraphics[width=1\textwidth]{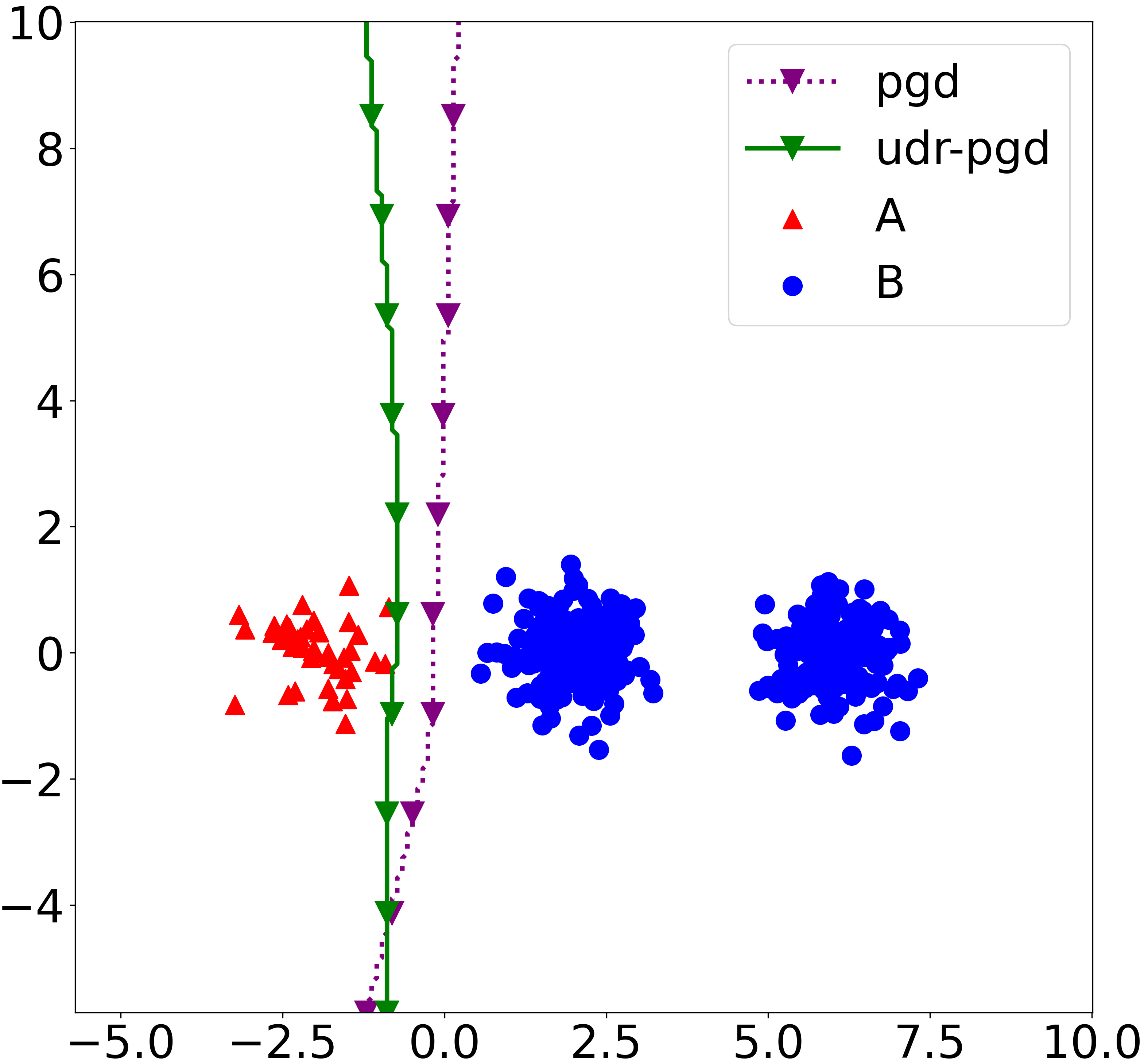}
\caption{Decision Boundary}
\label{fig:compare-db-1} \end{subfigure} \caption{(a)/(b): Trajectory of PGD and UDR-PGD adversarial examples. Each
trajectory includes 20 intermediate steps. For better visualization,
we do not use random initialization. The model is the natural training
at epoch 1. (c) The final decision boundary comparison.}
\end{figure}

We conduct an experiment on a synthetic dataset with a simple MLP
model to visualize the benefit of our UDR framework over the standard
AT methods, by taking UDR-PGD and PGD-AT as examples. The synthetic
dataset consists of three clusters A, B1, B2 where A, B are two classes
as shown in Figure \ref{fig:compare-db-1}. The data points are sampled
from normal distributions, i.e., $A\sim\mathcal{N}\left((-2,0),\Sigma\right),B1\sim\mathcal{N}\left((2,0),\Sigma\right)$
and $B2\sim\mathcal{N}\left((6,0),\Sigma\right)$ where $\Sigma=0.5*I$
with $I$ is the identity matrix. There are total 10k training samples
and 2k testing samples with densities of three clusters are 10\%,
50\% and 40\%, respectively. We use a simple model of 4 Fully-Connected
(FC) layers as follows: Input --> ReLU(FC(10)) --> ReLU(FC(10))
--> ReLU(FC(10)) --> Softmax(FC(2)), where FC(k) represents for
FC with k hidden units. We use Adam optimizer with learning rate 1e-3
and train with 30 epochs. We use $\{k=20,\epsilon=1.0,\eta=0.1\}$
for adversarial training (either PGD-AT or UDR-PGD) and PGD attack
with $\{k=200,\epsilon=2.0,\eta=0.1\}$ for evaluation.

It is a worth noting that while the distance between clusters is 2,
we limit the perturbation $\epsilon=1$ for the adversarial training
to show the advantage on the flexibility of the soft-ball projection
on the same/limited perturbation budget. Intuitively, cluster A has
the lowest density (10\%), therefore, the ideal decision boundary
should be surrounded cluster A which sacrifices the robustness of
the cluster A but increases the overall robustness eventually.

\paragraph{Comparison between UDR-PGD and PGD-AT.}

First, we visualize the trajectory of adversarial example from PGD
and our UDR-PGD as in Figures \ref{fig:pgd-traject},\ref{fig:udr-pgd}
to compare behaviors of two adversaries on the same pre-trained model.
It can be seen that: (i) the PGD's adversarial examples and ours are
pushed toward the lower confident region to maximize the prediction
loss $g_{\theta}(x',x,y)$; (ii) however, while the adversarial examples
of PGD are limited on the hard-projection ball, our adversarial examples
have more flexibility. Specifically, those are close to the decision
boundary (cluster A, B1) can go further, while those are distant to
the decision boundary (cluster B2) stay close to the original input.
This flexibility helps the adversarial examples reach better local
optimum of the prediction loss, hence, benefits the adversarial training.
Consequently, as shown in Figure \ref{fig:compare-db-1} the final
decision boundary of our UDR-PGD is closer to the ideal decision boundary
than that of PGD-AT, hence, achieving a better robustness. Quantitative
result shows that the robust accuracy of our UDR-PGD is 82.6\%, while
that of PGD-AT is 74.5\% with the same PGD attack $\{k=200,\epsilon=2.0,\eta=0.1\}$.

\paragraph{Comparison among UDR-PGD settings.}

\begin{figure}[ht!]
\centering \begin{subfigure}[b]{0.25\textwidth} \centering \includegraphics[width=1\textwidth]{images/toy2d_adv_trajectory_attack=udr_attack_random_init=False_alpha=1\lyxdot 0_lamda=0\lyxdot 1_num_steps=20_scale=0\lyxdot 2}
\caption{$\lambda=0.1,\tau=1.0$}
\label{fig:udr-traject1} \end{subfigure} \begin{subfigure}[b]{0.25\textwidth}
\centering \includegraphics[width=1\textwidth]{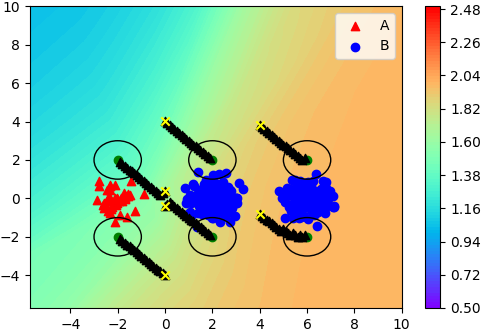}
\caption{$\lambda=0.01,\tau=1.0$}
\label{fig:udr-traject2} \end{subfigure} \begin{subfigure}[b]{0.25\textwidth}
\centering \includegraphics[width=1\textwidth]{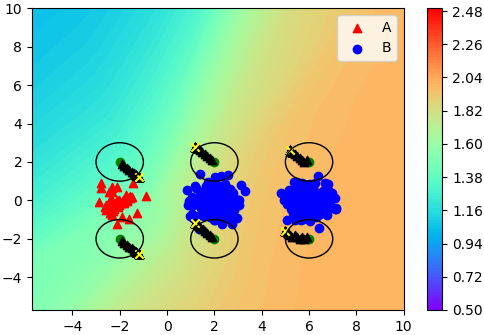}
\caption{$\lambda=0.01,\tau=0.01$}
\label{fig:udr-traject3} \end{subfigure} \begin{subfigure}[b]{0.20\textwidth}
\centering \includegraphics[width=1\textwidth]{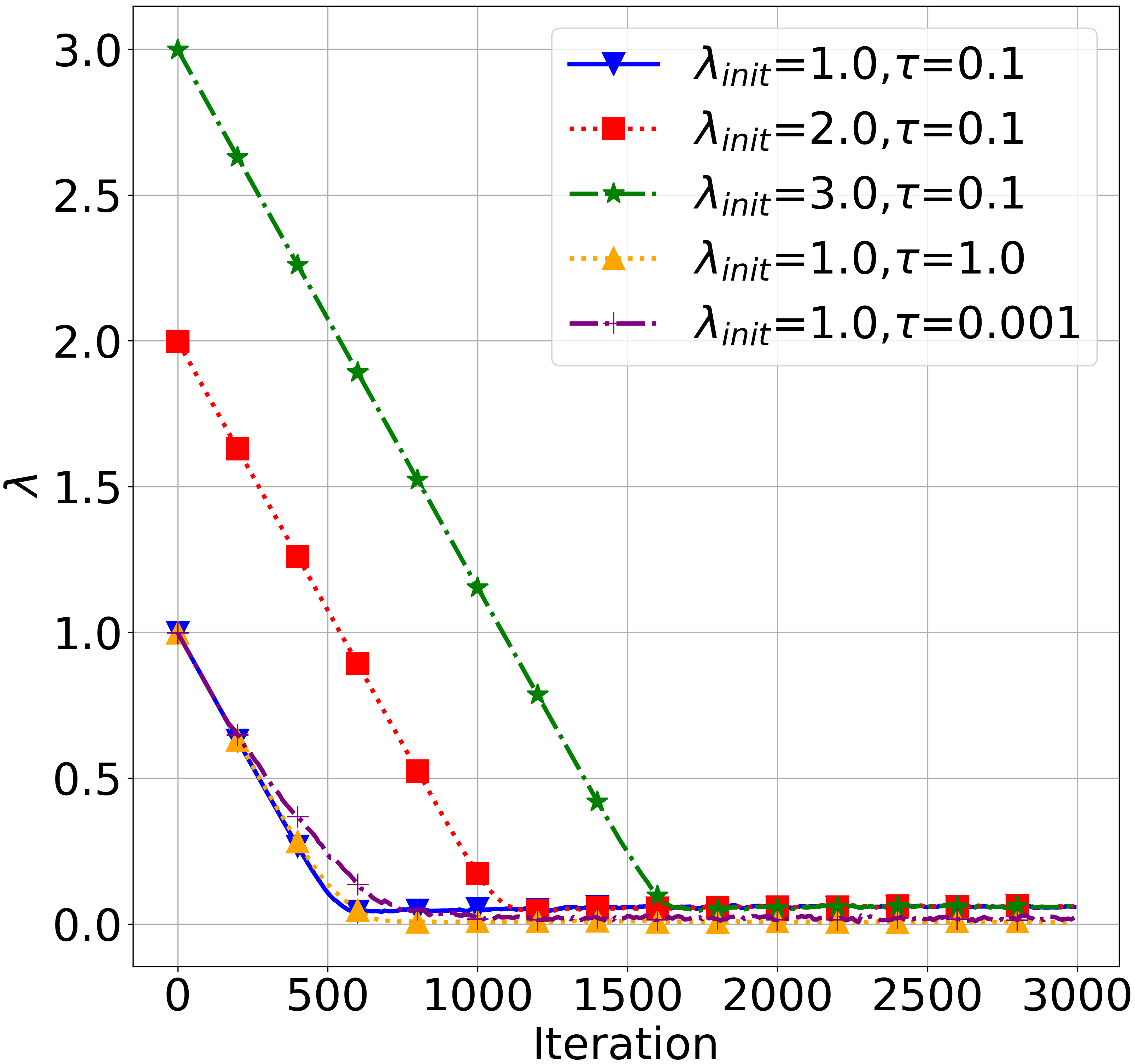}
\caption{Changing of $\lambda$}
\label{fig:lamda} \end{subfigure} \caption{(a)/(b)/(c): Trajectory of UDR-PGD adversarial examples with different
settings. Each trajectory includes 20 intermediate steps. For better
visualization, we do not use random initialization. The model is the
natural training at epoch 1. (d) The changing of parameter $\lambda$.}
\label{fig:trajectory} 
\end{figure}

Here we would like to provide more understanding about our framework
through the experiment with PGD-AT as shown in Figure \ref{fig:trajectory}.
First, we compare the trajectories of the adversarial examples of
UDR-PGD with different $\lambda$ as shown in Figures \ref{fig:udr-traject1},\ref{fig:udr-traject2}.
It can be seen that the crafted adversarial examples stay closer to
their benign counterparts when $\lambda$ becomes higher (i.e., $\lambda=0.1$
in Figure \ref{fig:udr-traject1}). In contrast, the soft-projection
ball is extended when $\lambda$ becomes smaller (i.e., $\lambda=0.01$
in Figure \ref{fig:udr-traject2}). On the other hand, with the same
$\lambda$ but smaller $\tau$as shown in Figure \ref{fig:udr-traject3},
the soft-ball projection is more identical to the hard ball projection
as shown in Figure \ref{fig:pgd-traject}. These behaviors concur
with the theoretical expectation as discussed in Section 4.1 in the
main paper.

Figure \ref{fig:lamda} shows the learning progress of parameter $\lambda$.
It can be observed that (i) the $\lambda$ converges to 0 regardless
of its initialization value and (ii) the convergence rate of $\lambda$
depends on the parameter $\tau$ (i.e., smaller $\tau$ slower convergence).
We choose $\tau=2\eta$ for the experiments on real-world image datasets.

\paragraph{Further results of soft-ball projection.}

\begin{figure}[ht!]
    \centering
    \includegraphics[width=0.5\textwidth]{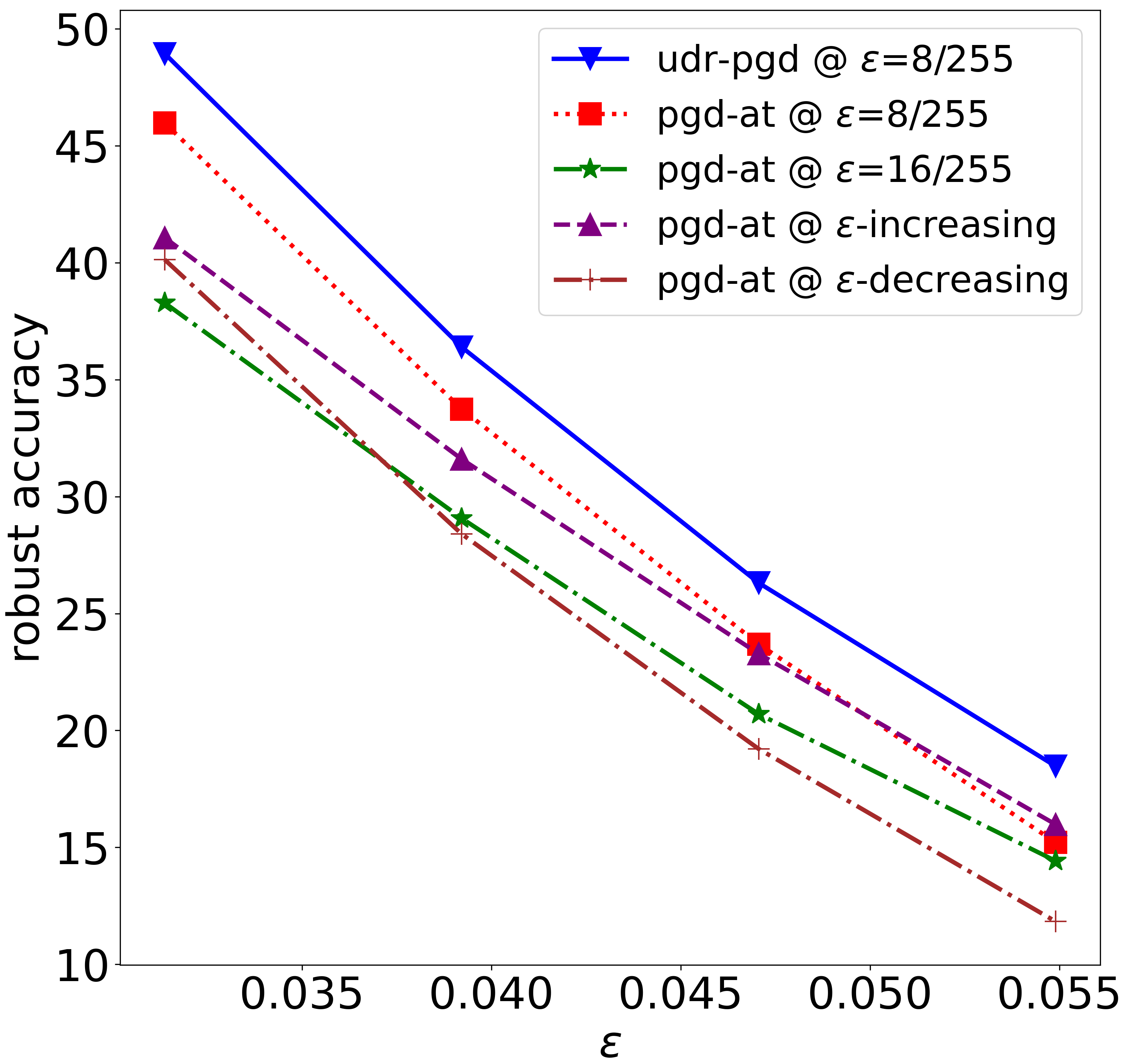}
    \caption{Hard/soft-ball projections}
    \label{fig:ablation-pgd}
\end{figure}

In Figure \ref{fig:ablation-pgd}, we compare our UDR-PGD with the soft-ball
projection to PGD-AT with the hard-ball projection with different
settings against the PGD attack on CIFAR10. For PGD-AT, we use the
following three ad-hoc strategies for $\epsilon$: 1) Fixing $\epsilon=8/255$;
2) Fixing $\epsilon=16/255$; 3) Gradually increasing/decreasing $\epsilon$
from 8/255 to 16/255 (Refer to Appendix \ref{sec:Exp-Setting} for
details). It can be seen that it is hard to find an effective strategy
of the perturbation boundary of the hard-ball projection for PGD-AT,
which can outperform ours. This demonstrates the benefit of our soft-project
operation. 

\section{More results and analysis\label{sec:supp-exp} }

\paragraph{Further results with C\&W (L2) attack.} 
We enrich the comprehensiveness of the experiments by further evaluating the defense methods with C\&W (L2) attack \citep{carlini2017towards} which is a very strong optimization based attack. 
The experiment has been conducted on the CIFAR10 dataset with WideResNet architecture. 
The hyper-parameters are $c\in \{0.5,0.7,1.0\}, kappa=0, steps=1000, lr=0.01$ where $kappa$ is the 
confidence coefficient and $c$ is box-constraint coefficient.\footnote{We use the implementation 
from https://github.com/Harry24k/adversarial-attacks-pytorch} 
As shown in Table \ref{exp:CWL2}, our distributional robustness version significantly outperform 
the standard ones in term of robust accuracy. For example, against C\&W (c=0.5) attack, 
the robust accuracy gap between UDR-PGD and PGD-AT is $6\%$  while that for UDR-AWP-AT and 
AWP-AT is around $5\%$. The average improvement of robust accuracies against different 
levels of attack strengths is around $5\%$. This result strongly emphasizes the 
contribution of our distributional robustness and the soft-ball projection over the standard adversarial training.

\begin{table}
    \caption{Robustness evaluation against C\&W attack with
    WRN-34-10 on the full test set of the CIFAR10 dataset (10K test images). $c$ is box-constraint coefficient. 
	({*}) Omit the cross-entropy loss of natural images.}
    \centering
    \label{exp:CWL2}
	\begin{tabular}{lccccc}
		\hline 
		& Nat & $c=0.5$ & $c=0.7$ & $c=1.0$ & Avg-Gap\tabularnewline
		\hline 
		PGD-AT{*} & 84.93 & 40.85 & 25.90 & 12.95 & - \tabularnewline
		UDR-PGD{*} & 84.60 & \textbf{47.31} & \textbf{31.58} & \textbf{16.57} & 5.25\tabularnewline
		\hline 
		TRADES & 85.70 & 47.65 & 34.30 & 21.03 & -\tabularnewline
		UDR-TRADES & 84.93 & \textbf{49.14} & \textbf{36.33} & \textbf{23.28} & 1.92\tabularnewline
		\hline 
		AWP-AT & 85.57 & 49.91 & 34.31 & 18.97 & - \tabularnewline
		UDR-AWP-AT & 85.51 & \textbf{54.44} & \textbf{39.86} & \textbf{23.61} & 4.91\tabularnewline
		\hline 		
	\end{tabular}  
\end{table}

\paragraph{Experimental results of WRM \citep{sinha2017certifying}.} 
The performance of WRM highly depends on the Lagrange dual parameter $\gamma$ (or $\epsilon=0.5/\gamma$ in their 
implementation\footnote{https://github.com/duchi-lab/certifiable-distributional-robustness/blob/master/attacks\_tf.py}), which 
controls the robustness level. As mentioned in their paper, with large $\gamma$, the method is less robust 
but more tractable. Generally, decreasing $\gamma$ will reduce the natural accuracy but increase the robustness 
of the model as shown in Table \ref{exp:wrm-study}. We obtained the best performance on MNIST 
with $\gamma=0.05$ (CNN), while on CIFAR10 and CIFAR100 with $\gamma=0.5$ (ResNet18). The best results 
with three benchmark datasets have been reported as in Table \ref{tab:whitebox-with-wrm} (recall results 
from Table \ref{tab:whitebox}). It is a worth mentioning that while we could obtain a similar performance 
as reported Sinha et al. (2017) on the MNIST dataset with their architecture (3 Convolution layers + 1 FC layer), 
however, WRM seems much less effective with larger architectures.

\begin{table} [H]
    \caption{Result of WRM with different $\epsilon=0.5/\gamma$ on the CIFAR10 dataset.}
    \label{exp:wrm-study}
    \centering \resizebox{0.3\linewidth}{!}{ %
    \begin{tabular}{@{}ccccc@{}}
    \toprule 
     & Nat  & PGD  & AA  & B\&B  \tabularnewline
    $\epsilon=0.1$  & 90.9  & 15.3  & 13.7  & 15.8  \tabularnewline
    $\epsilon=0.5$  & 86.7  & 33.9  & 32.6  & 35.4  \tabularnewline
    $\epsilon=1.0$  & 83.7  & 40.9  & 39.8  & 41.4  \tabularnewline
    $\epsilon=2.0$  & 79.4  & 45.4  & 43.6  & 45.5  \tabularnewline
    $\epsilon=5.0$  & 71.6  & 47.5  & 45.2  & 46.2  \tabularnewline
    $\epsilon=10.0$  & 65.0  & 46.6  & 43.4  & 44.4  \tabularnewline
    \bottomrule
    \end{tabular}}
    \end{table}

\begin{table}
    \caption{Comparisons of natural classification accuracy (Nat) and adversarial accuracies against different attacks. Recall results from Table \ref{tab:whitebox} with additional results of WRM. Best scores are highlighted in boldface.}
    \label{tab:whitebox-with-wrm}
    \centering \resizebox{0.9\linewidth}{!}{ %
    \begin{tabular}{@{}ccccccccccccccc@{}}
    \toprule 
     & \multicolumn{4}{c}{MNIST} &  & \multicolumn{4}{c}{CIFAR10} &  & \multicolumn{4}{c}{CIFAR100}\tabularnewline
    \midrule 
     & Nat  & PGD  & AA  & B\&B  &  & Nat  & PGD  & AA  & B\&B  &  & Nat  & PGD  & AA  & B\&B\tabularnewline
    WRM  & 91.8  & 27.1  & 4.5  & 8.2  &  & 83.7  & 40.9  & 39.8  & 41.4  &  & 56.6  & 24.7  & 21.3  & 22.9 \tabularnewline
    PGD-AT  & 99.4  & 94.0  & 88.9  & 91.3  &  & \textbf{86.4}  & 46.0  & 42.5  & 44.2  &  & 72.4  & 41.7  & 39.3  & 39.6\tabularnewline
    UDR-PGD  & \textbf{99.5}  & \textbf{94.3}  & \textbf{90.0}  & \textbf{91.4}  &  & \textbf{86.4}  & \textbf{48.9}  & \textbf{44.8}  & \textbf{46.0}  &  & \textbf{73.5}  & \textbf{45.1}  & \textbf{41.9}  & \textbf{42.3}\tabularnewline
    \bottomrule
    \end{tabular}}
    \end{table}
    
\paragraph{Further results of whitebox attacks with varied $\epsilon$.}

\begin{figure}[ht!]
\centering \begin{subfigure}[b]{0.3\textwidth} \centering \includegraphics[width=1\textwidth]{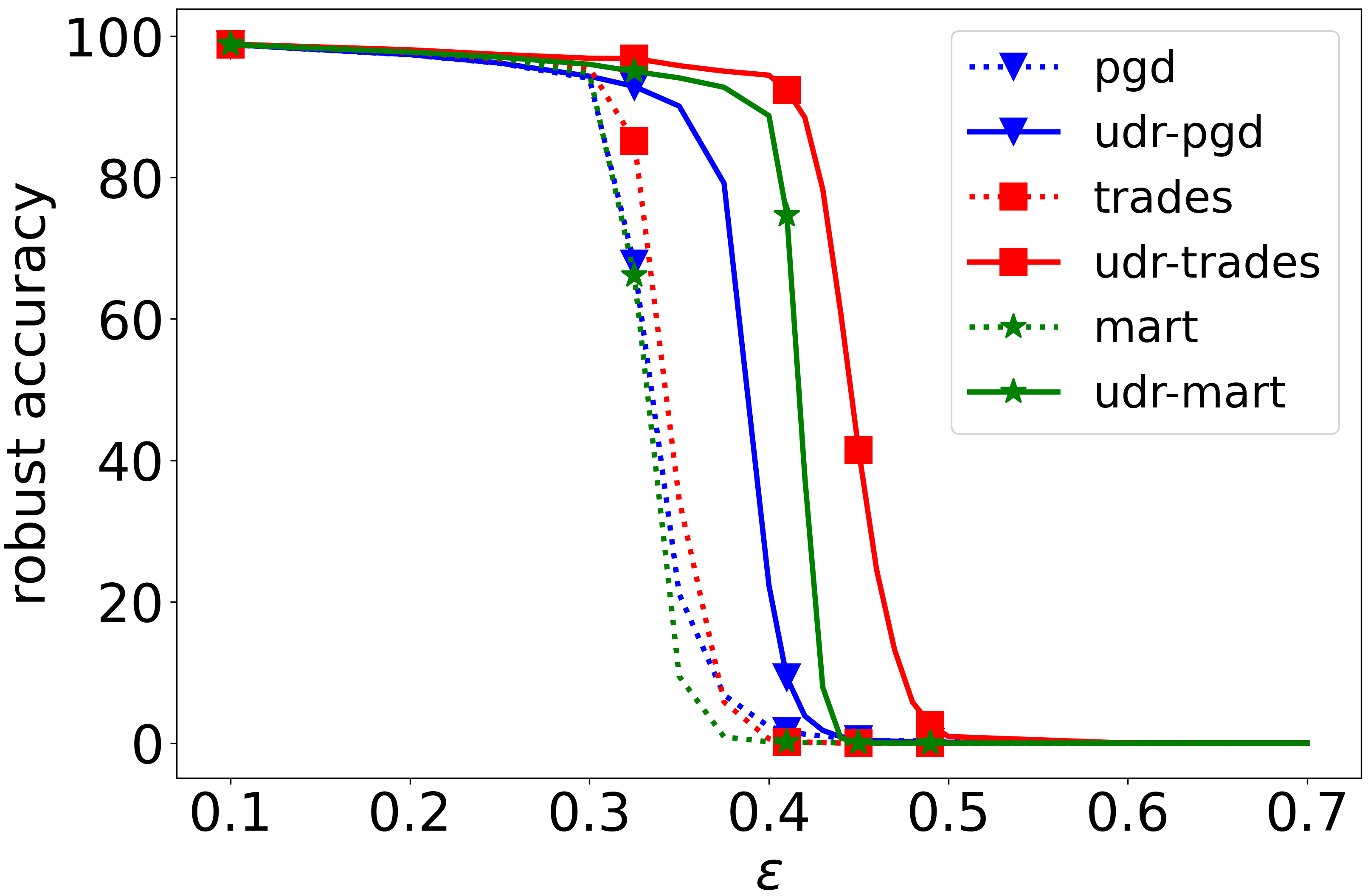}
\caption{MNIST}
\end{subfigure} \begin{subfigure}[b]{0.3\textwidth} \centering
\includegraphics[width=1\textwidth]{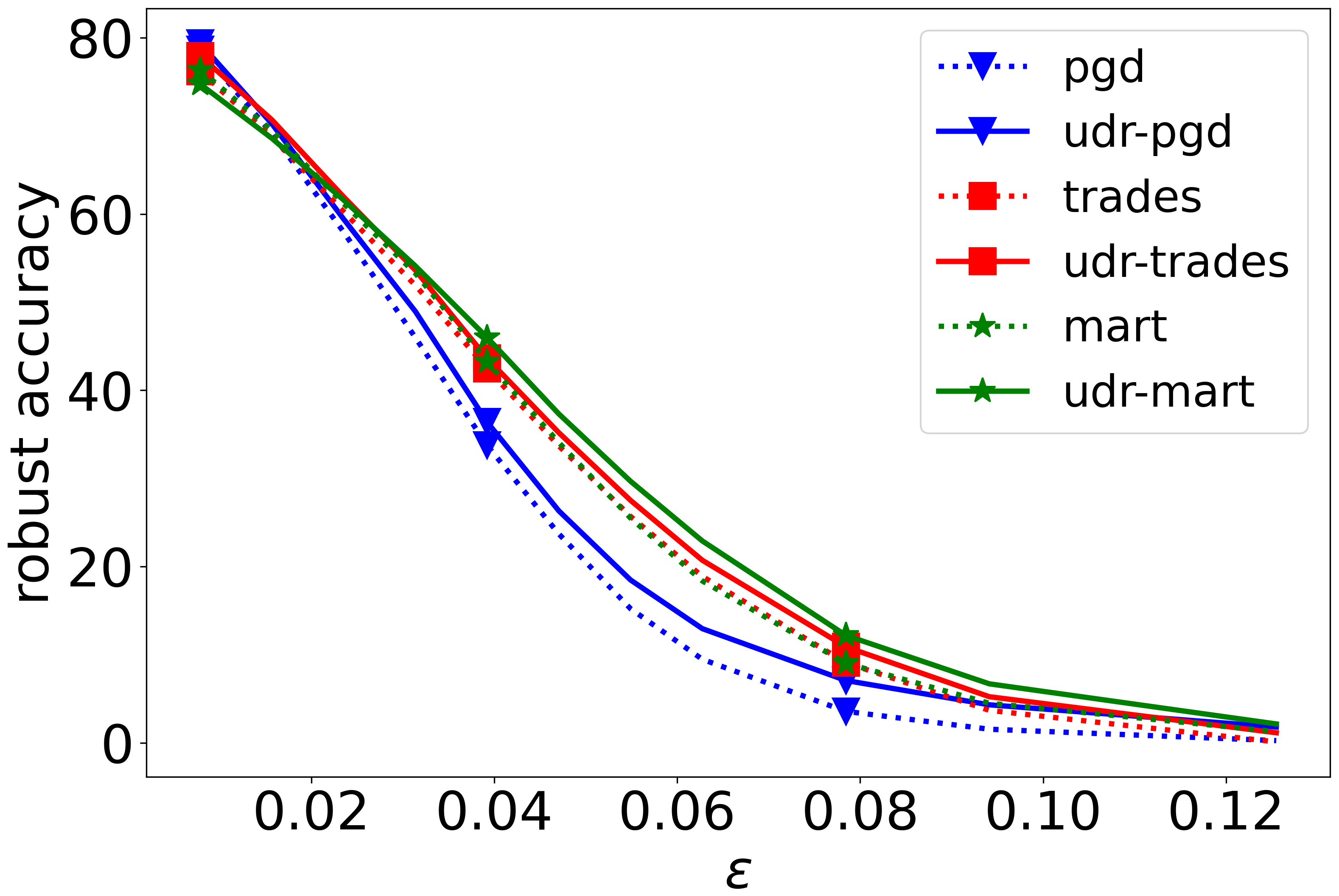}
\caption{CIFAR10}
\end{subfigure} \begin{subfigure}[b]{0.3\textwidth} \centering
\includegraphics[width=1\textwidth]{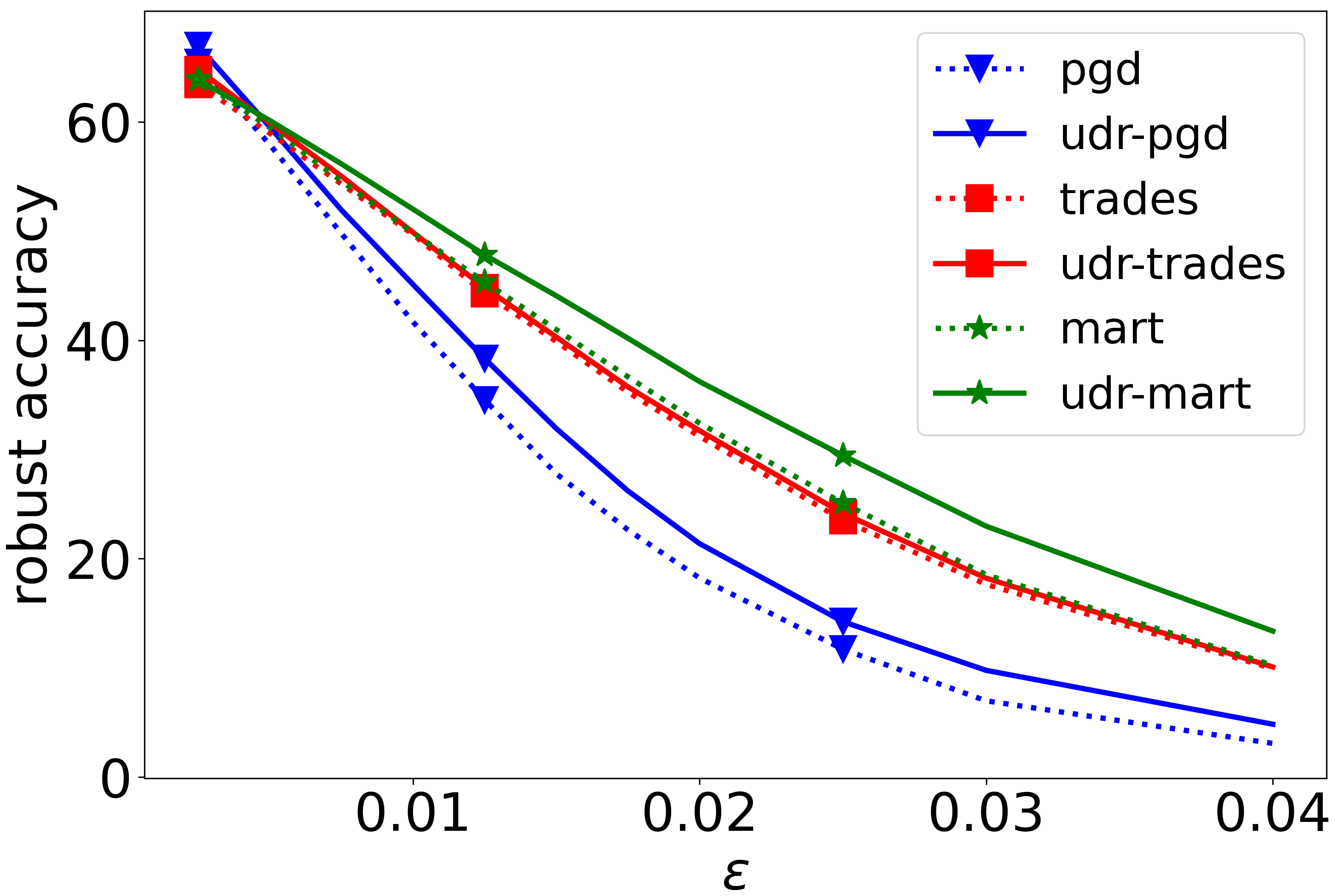}
\caption{CIFAR100}
\end{subfigure} \caption{Robustness evaluation against multiple attack strengths.}
\label{fig:supp-multi-eps} 
\end{figure}

Here we would like to provide more results on defending against whitebox
attacks with a bigger range of $\epsilon$ as shown in Figure \ref{fig:supp-multi-eps}.
It can be seen that in a wide range of attack strengths our DR methods
consistently outperform their AT counterparts.

\paragraph{The convergence of the algorithm.}

\begin{figure}[ht!]
\centering \begin{subfigure}[b]{0.3\textwidth} \centering \includegraphics[width=1\textwidth]{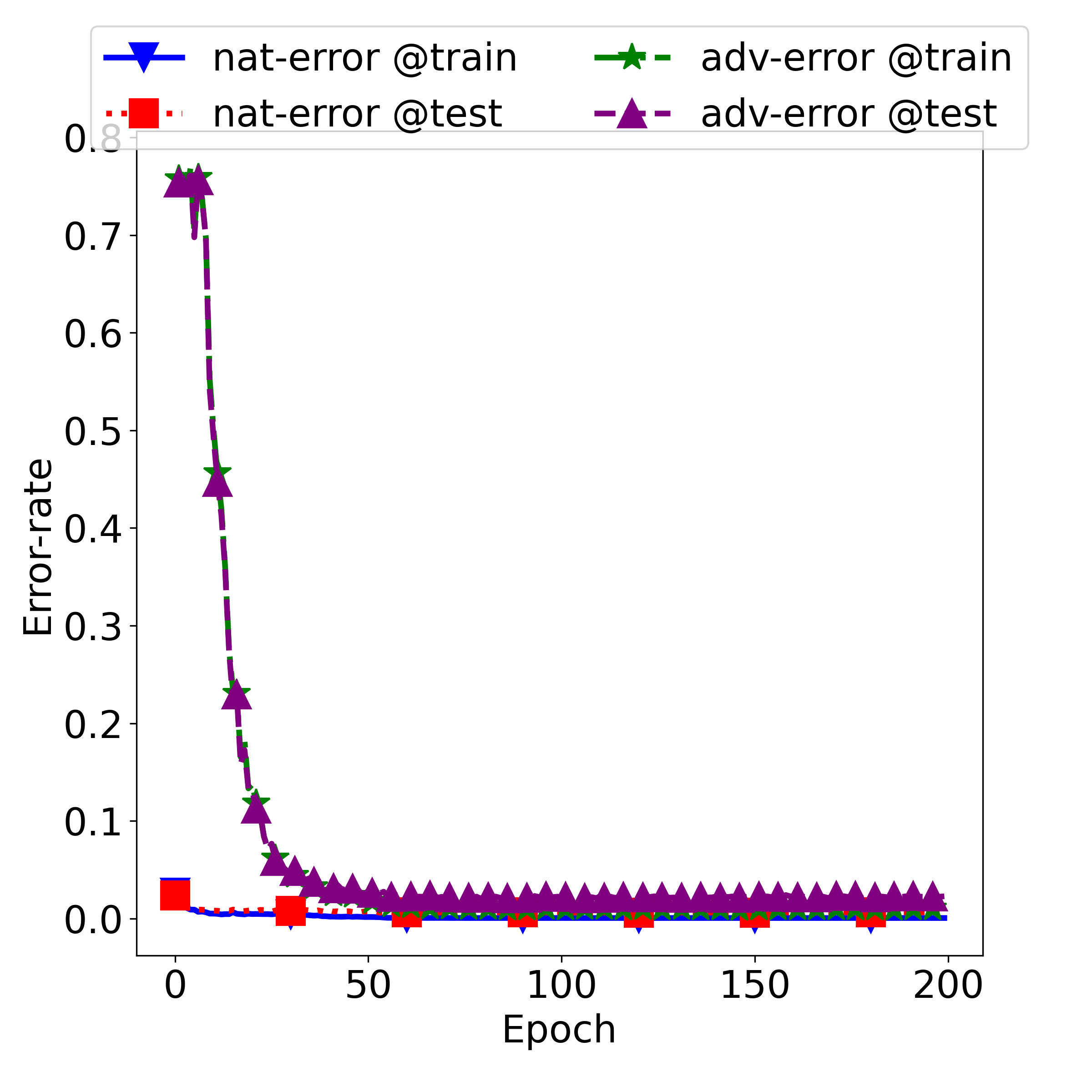}
\caption{MNIST}
\end{subfigure} \begin{subfigure}[b]{0.3\textwidth} \centering
\includegraphics[width=1\textwidth]{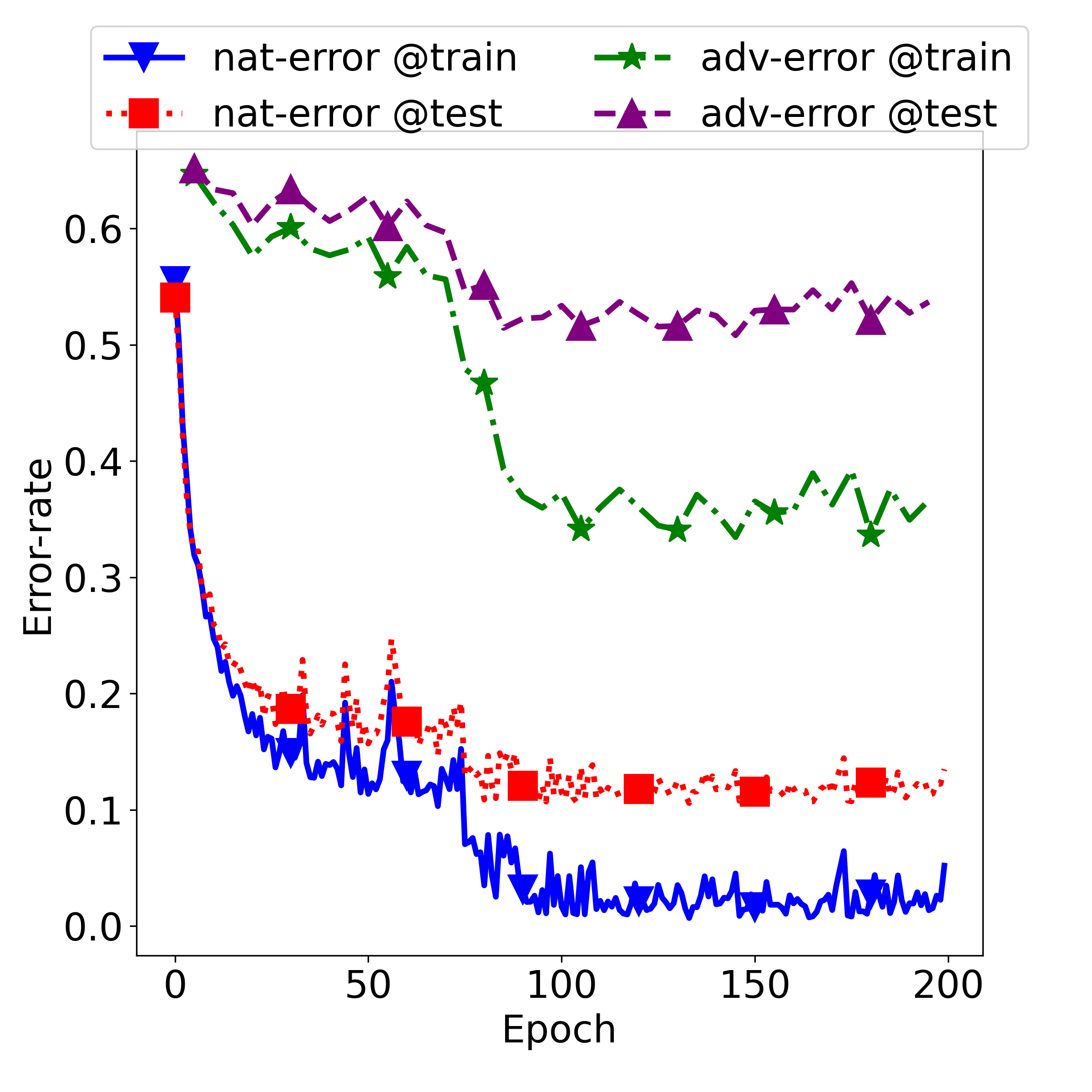}
\caption{CIFAR10}
\end{subfigure} \begin{subfigure}[b]{0.3\textwidth} \centering
\includegraphics[width=1\textwidth]{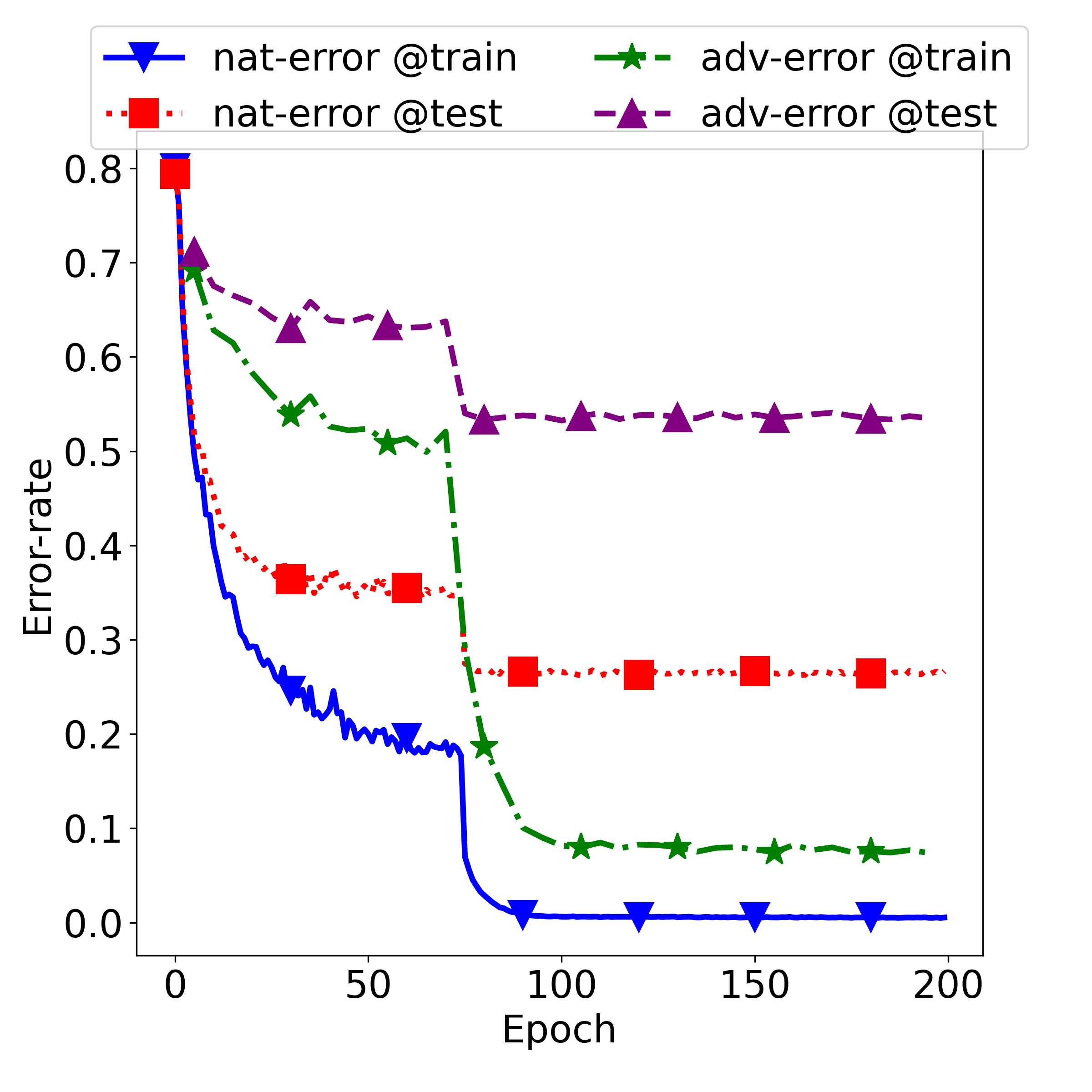}
\caption{CIFAR100}
\end{subfigure} \caption{Training progress of our UDR-PGD on different datasets, evaluating
on the full training set (e.g., 50k images) and the full testing set
(e.g., 10k images). Robust accuracy is against PGD attack with $k=20$.}
\label{fig:training-progress} 
\end{figure}

During the training, we observed that while adversarial examples distribute
inside/outside the hard ball $\epsilon$ differently (i.e., as shown
in Figure \ref{fig:udr-pgd} ), but generally the average distance
to original input is less than $\epsilon$. Therefore, according to
the update formulation in Eq. (19), $\lambda$ tends to decrease to
0 and eventually is stable at 0 because of very small learning rate
as shown in Figure \ref{fig:lamda}. In addition, we visualize the
training progress as shown in Figure \ref{fig:training-progress}
to show the convergence of our method. It can be seen that, the error-rate
reduces over training progress and converges at the end of the training
progress.

\paragraph{Further experiment result on CIFAR100.}

We would like to provide additional experiment result on CIFAR100
dataset such that all defenses are adversarially trained with $\epsilon=\frac{8}{255}$.
Our UDR-PGD outperforms PGD 3.7\% at $\epsilon=\frac{8}{255}$ and
2.3\% on average, while our UDR-TRADES and UDR-MART outperform their
counterparts by around 0.5\% and 0.7\%, respectively. It is worth
noting that, in our experiment, MART is quite sensitive with changes
of (MART’s natural accuracy drops to a lower performance than that
of TRADES); that might explain the lower gap between UDR-MART and
MART with the new $\epsilon$.

\begin{table}

\caption{Robustness evaluation on CIFAR100 dataset. The last column “Avg” represents
the average gap of robust accuracy between our methods and their standard
AT counterparts.}

\begin{centering}
\begin{tabular}{lcccccccc}
\hline 
 & Nat & $\frac{8}{255}$ & $\frac{10}{255}$ & $\frac{12}{255}$ & $\frac{14}{255}$ & $\frac{16}{255}$ & $\frac{20}{255}$ & Avg\tabularnewline
\hline 
PGD-AT & 63.7 & 22.8 & 16.1 & 11.4 & 7.8 & 5.1 & 2.4 & -\tabularnewline
UDR-PGD & 64.5 & 26.5 & 18.9 & 13.7 & 9.8 & 7.0 & 3.5 & 2.30\tabularnewline
\hline 
TRADES & 60.2 & 30.3 & 24.5 & 18.8 & 14.8 & 11.5 & 6.7 & -\tabularnewline
UDR-TRADES & 60.1 & 30.8 & 25.1 & 19.3 & 15.5 & 12.2 & 7.5 & 0.52\tabularnewline
\hline 
MART & 54.1 & 32.0 & 26.8 & 21.9 & 17.4 & 13.8 & 7.6 & -\tabularnewline
UDR-MART & 54.4 & 32.3 & 27.4 & 22.5 & 18.4 & 14.4 & 8.5 & 0.67\tabularnewline
\hline 
\end{tabular}
\par\end{centering}
\end{table}

\section{Choosing The Cost Function\label{sec:chose-cost-func}}

In this section, we provide the technical details of our learning
algorithm in Section 4 in the main paper, especially, the important
of choosing cost function $\hat{c}(x,x')$. Given the current model
$\theta$ and the parameter $\lambda$, we find the adversarial examples
by solving: 
\[
x^{a}=\text{argmax}_{x'}\left\{ g_{\theta}(x',x,y)-\lambda\hat{c}_{\mathcal{X}}\left(x',x\right)\right\} 
\]

We employ multiple gradient ascent update steps without projecting
onto the hard ball $B_{\epsilon}$. Specifically, the updated adversarial
at step $t+1$ as follows:

\[
x^{t+1}=x^{t}+\eta\,\left(\nabla_{x'}g_{\theta}(x',x,y)-\lambda\,\nabla_{x'}\hat{c}_{\mathcal{X}}\left(x',x\right)\right)
\]

Given the smoothed cost function as in Equation (19), the updating
step is as follows:

\[
x^{t+1}=\begin{cases}
x^{t}+\eta\,\left(\nabla_{x'}g_{\theta}(x',x,y)-\lambda\,\nabla_{x'}c_{\mathcal{X}}\left(x',x\right)\right), & \text{if}\;c_{\mathcal{X}}\left(x',x\right)<\epsilon\\
x^{t}+\eta\,\left(\nabla_{x'}g_{\theta}(x',x,y)-\frac{\lambda}{\tau}\,\nabla_{x'}c_{\mathcal{X}}\left(x',x\right)\right), & \text{otherwise.}
\end{cases}
\]

It shows that, the pixels that are out-of-perturbation ball $B_{\epsilon}$
will be traced back with a longer step, depending on the parameter
$\tau$. We consider three popular distance functions of $c_{\mathcal{X}}\left(x',x\right)$
with their gradient as Table \ref{tab:dis-funcs}. It is worth noting
that, while the norm $L_{1},L_{2}$ have gradient in all pixels, the
$L_{\infty}$ has gradient in only one pixel per image. It means that,
when using $L_{\infty}$ norm as the cost function $c_{\mathcal{X}}(x,x')$,
only single pixel has been traced back at each iteration. In contrast,
using $L_{2}$ will project all pixels toward the original input $x$
with the step size of each. As in the discussion in Section \ref{sec:supp-exp},
only small part of an MNIST image contributes to the prediction, while
in contrast, most of pixels of a CIFAR10 image affect to the prediction.
Based on this observation, we use the $L_{\infty}$ for the MNIST
dataset and $L_{2}$ for the CIFAR10 dataset in the updating step.
However, the perturbation strength $\epsilon$ has been measured in
$L_{\infty}$, therefore, we still use $L_{\infty}$ in the Equation
(22) to update $\lambda$.

\begin{table}
\caption{Distance function and its gradient \label{tab:dis-funcs}}

\centering{}\resizebox{0.45\textwidth}{!}{\centering\setlength{\tabcolsep}{2pt}
\begin{tabular}{ccc}
\hline 
 & $c_{\mathcal{X}}(x,x')$  & $\nabla_{x'}c(x,x')$\tabularnewline
\hline 
$L_{1}$  & $\sum_{i=1}^{d}\left\Vert x_{i}-x{}_{i}^{'}\right\Vert $  & 1, $\forall i\in[1,d]$\tabularnewline
$L_{2}$  & $\frac{1}{2}\sum_{i=1}^{d}\left(x_{i}-x_{i}'\right)^{2}$  & $\sum_{i=1}^{d}(x'_{i}-x_{i})$\tabularnewline
$L_{\infty}$  & $\max_{i}\left\Vert x_{i}-x{}_{i}^{'}\right\Vert $  & $\begin{cases}
1,i=\text{argmax}_{i}\left\Vert x_{i}-x{}_{i}^{'}\right\Vert \\
0,\text{\text{otherwise}}
\end{cases}$\tabularnewline
\hline 
\end{tabular}} 
\end{table}

\begin{figure}[ht!]
\centering \begin{subfigure}[b]{0.22\textwidth} \centering \includegraphics[width=1\textwidth]{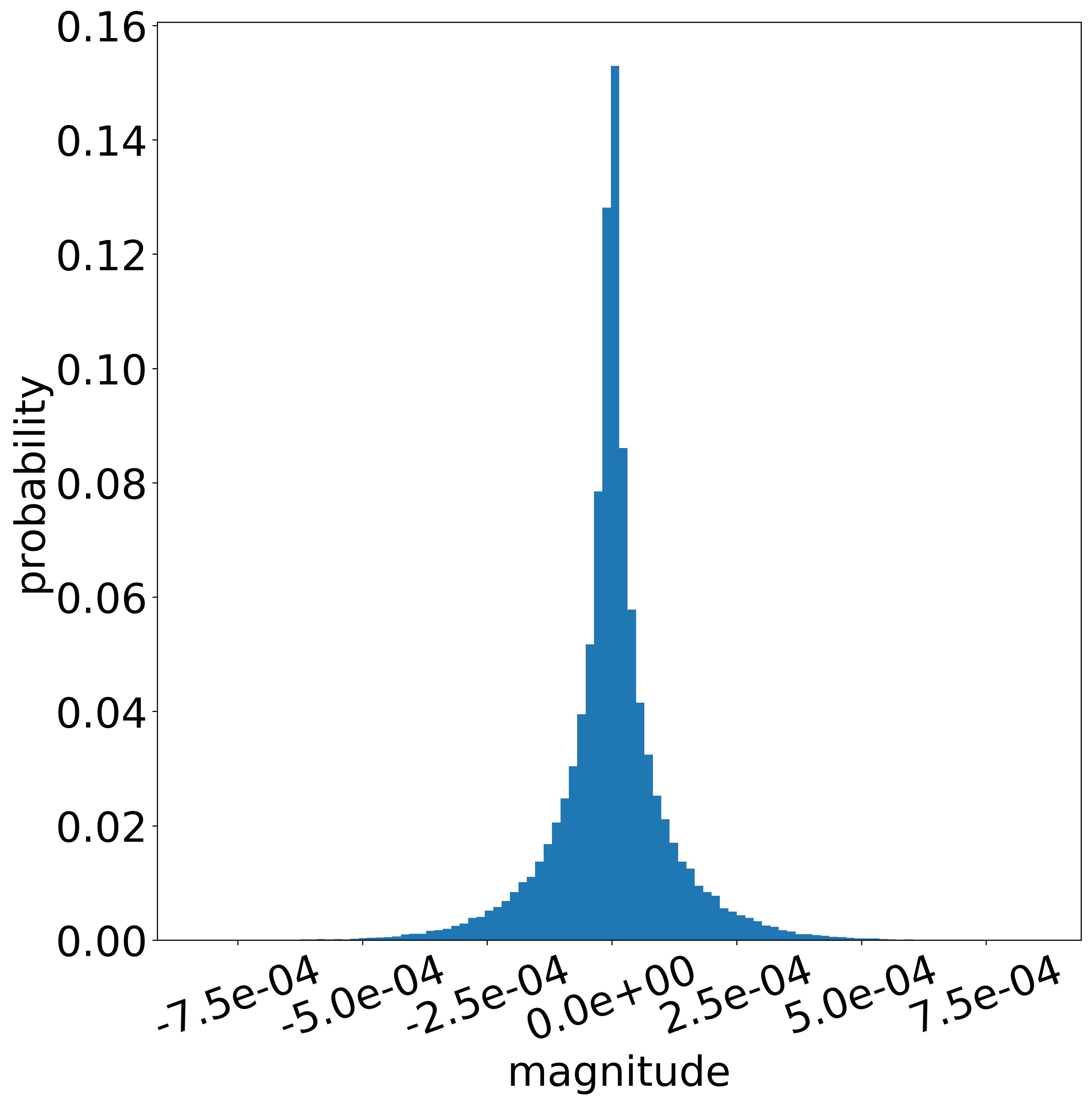}
\caption{CIFAR10, $grad1$}
\end{subfigure} \begin{subfigure}[b]{0.22\textwidth} \centering
\includegraphics[width=1\textwidth]{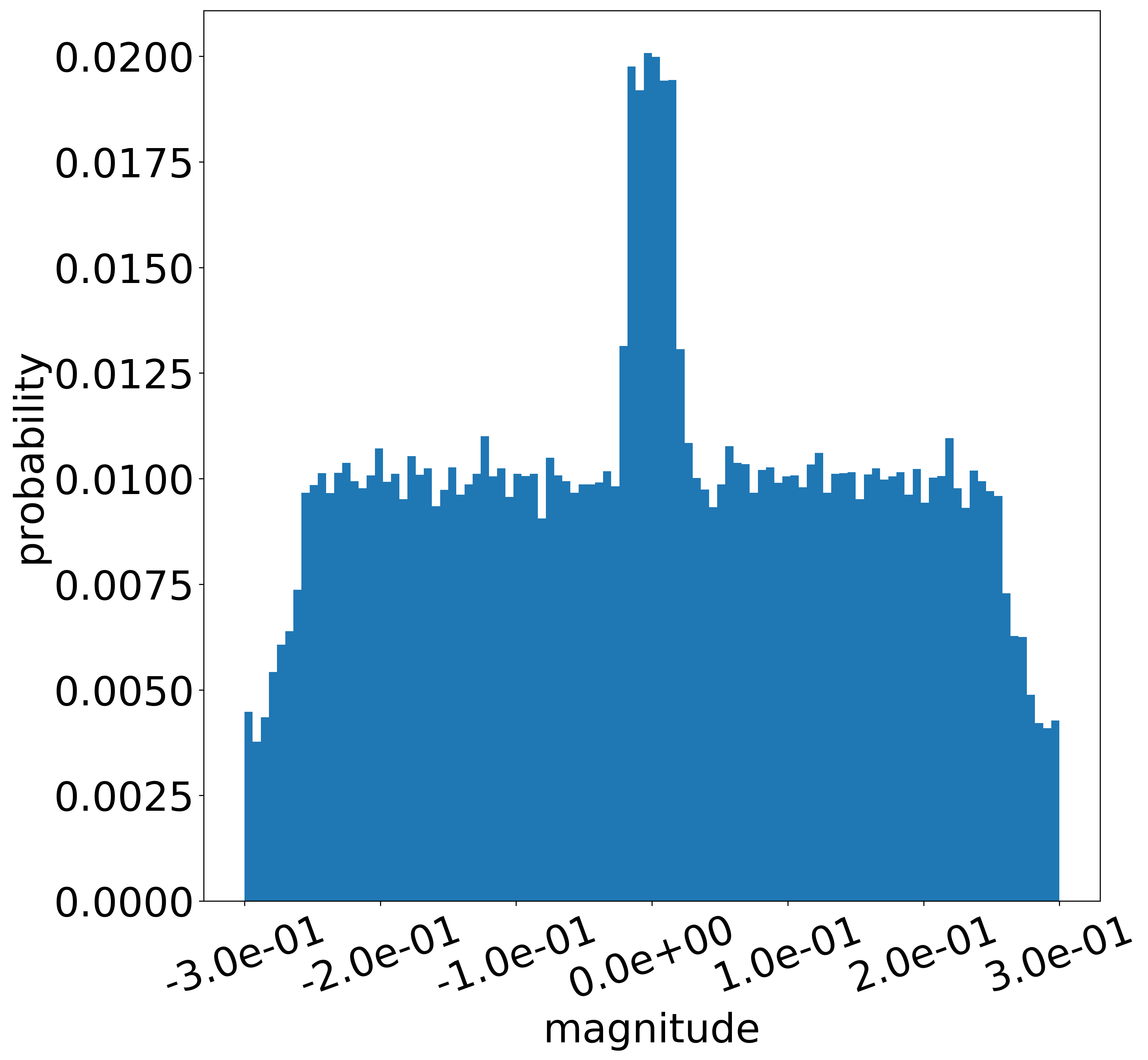}
\caption{MNIST, $grad2$}
\end{subfigure} \begin{subfigure}[b]{0.22\textwidth} \centering
\includegraphics[width=1\textwidth]{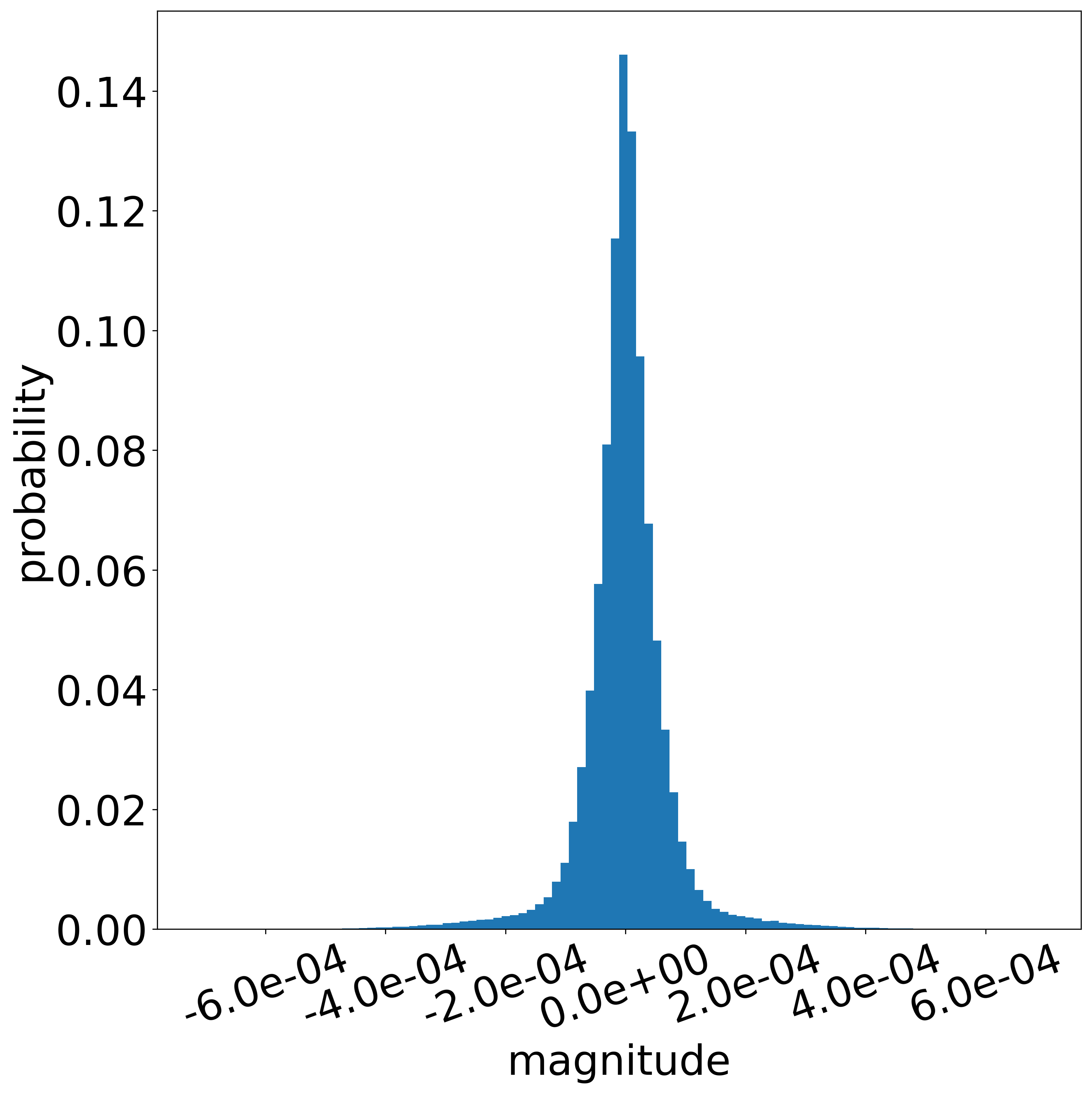}
\caption{CIFAR10, $grad1$}
\end{subfigure} \begin{subfigure}[b]{0.22\textwidth} \centering
\includegraphics[width=1\textwidth]{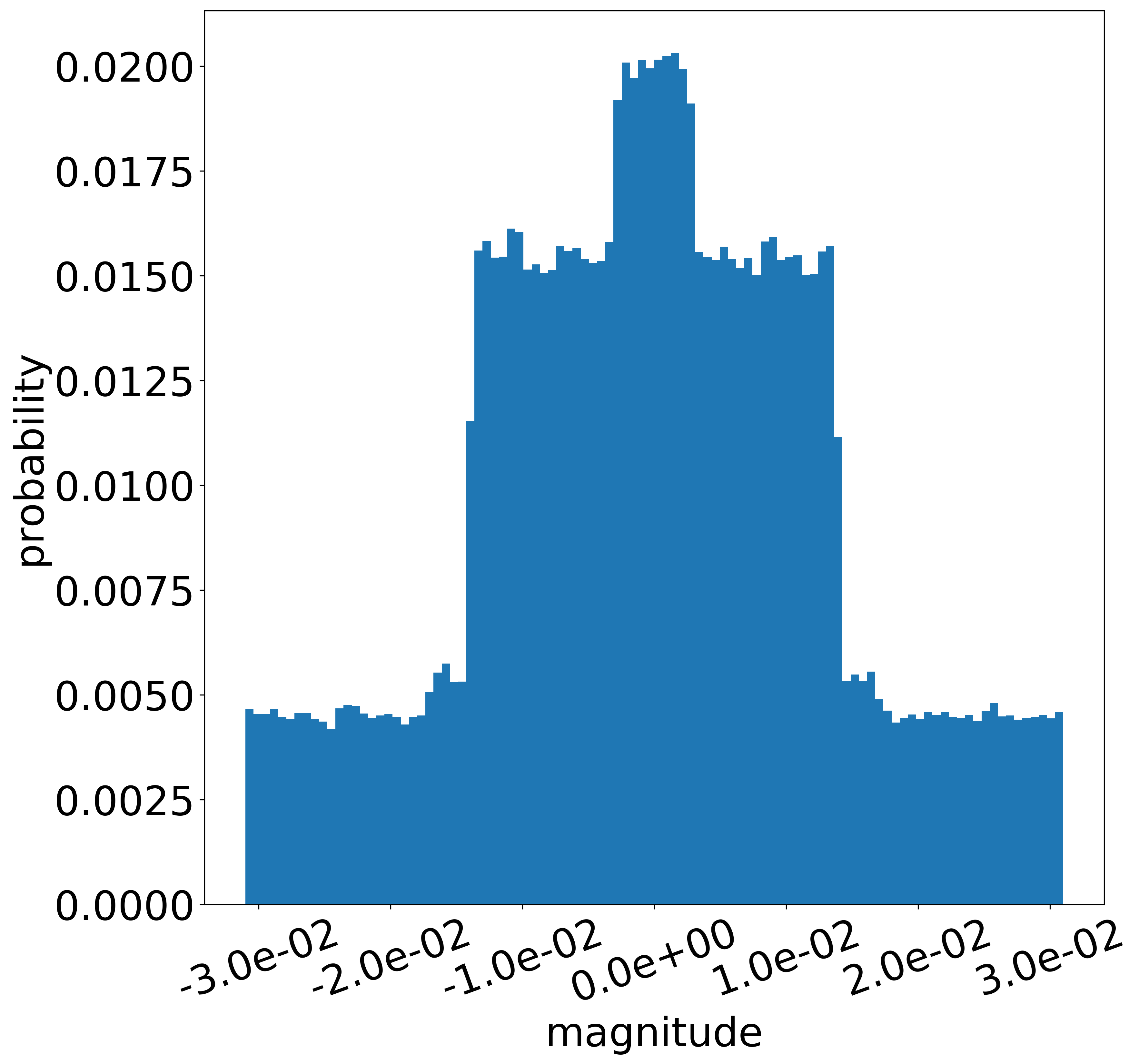}
\caption{CIFAR10, $grad2$}
\end{subfigure} \caption{Histogram of gradient strength of $grad1=\nabla_{x'}g_{\theta}(x',x,y)$
and $grad2=\nabla_{x'}\hat{c}_{\mathcal{X}}\left(x',x\right)$ on
MNIST and CIFAR10 dataset. We use $L_{2}$ norm for the cost function
$c_{\mathcal{X}}\left(x',x\right)$ , $\tau=\eta$ and $\lambda=1$}
\label{fig:his-grad} 
\end{figure}

We also visualize the histogram of gradient $\nabla_{x'}g_{\theta}(x',x,y)$
and $\nabla_{x'}\hat{c}_{\mathcal{X}}\left(x',x\right)$ as shown
in Figure \ref{fig:his-grad}. It can be seen that the strength of
gradient $grad1=\nabla_{x'}g_{\theta}(x',x,y)$ is much smaller than
$grad2=\nabla_{x'}\hat{c}_{\mathcal{X}}\left(x',x\right)$, for example,
on the MNIST dataset, $grad1\in[-5\times10^{-4},5\times10^{-4}]$
while $grad2\in[-0.3,0.3]$ which is 600 times larger. Therefore,
if using single update step, the gradient $\nabla_{x'}\hat{c}_{\mathcal{X}}\left(x',x\right)$
dominates the other and pulls the adversarial examples close to the
natural input. These adversarial examples are weaker and do not helps
to improve the robustness. Alternatively, we break single update step
for solving Equation (21) to two sub-steps as shown in Algorithm 1
to balance between push/pull steps. It also can be seen that the $grad2$
corresponds with the perturbation boundary $\epsilon$ and the step
size $\eta$. For example, on the MNIST dataset, $grad2$ has the
range from $[-0.3,0.3]$ and has the highest density around $[-0.01,0.01]$
where \{0.3, 0.01\} are the perturbation boundary and step size in
the experiment.

\end{document}